%% file: neurips_2026.tex
\documentclass{article}

\PassOptionsToPackage{numbers,sort&compress}{natbib}
\usepackage[main, final]{neurips_2026}
\usepackage{capt-of}

\usepackage[utf8]{inputenc} 
\usepackage[T1]{fontenc}    
\usepackage{hyperref}       
\usepackage{url}            
\usepackage{booktabs}       
\usepackage{amsfonts}       
\usepackage{nicefrac}       
\usepackage{microtype}      
\usepackage{xcolor}         
\usepackage{amsmath}
\usepackage{graphicx}
\usepackage{float}
\usepackage{stfloats}
\usepackage{afterpage}
\usepackage{placeins}
\usepackage{float}

\title{How Far Is Too Far? \\Object Recognition Declines Monotonically with Semantic Distance}

\author{%
  Domenic Bersch \\
  Department of Computer Science\\
  Goethe University\\
  Frankfurt, Germany \\
  \texttt{bersch@em.uni-frankfurt.de} \\
  \And
  Hai Van Tran\\
  Department of Computer Science\\
  Goethe University\\
  Frankfurt, Germany \\
  \texttt{s7671493@stud.uni-frankfurt.de} \\
  \And
  Gemma Roig\\
  Department of Computer Science\\
  Goethe University\\
  Frankfurt, Germany \\
  \texttt{roignoguera@em.uni-frankfurt.de} \\
}

\begin{document}

\maketitle

\begin{abstract}
Vision models are sensitive to object--background context, yet it remains unclear how recognition changes as the semantic distance between objects and backgrounds increases. We introduce ImageNet-OOC1k, a benchmark enabling graded control of object--background relationships along a continuous consensus visual-semantic dissimilarity axis. This formulation reveals a highly consistent monotonic decline in recognition accuracy across 35 models spanning convolutional neural networks, Vision Transformers, and vision--language models. Internal analyses of Vision Transformers reveal patterns consistent with aggregation failure, where object evidence is present at the patch level but not consolidated into the final prediction. Using this characterization as a diagnostic probe, a simple spatial re-ranking strategy recovers a subset of these errors and improves in-sample accuracy by up to 5.7 percentage points. These results establish context sensitivity as a continuous and predictable property of modern vision systems, with Vision Transformer analyses suggesting that aggregation limitations may contribute to some failure types even when object evidence remains available.
\end{abstract}

\begingroup
\setlength{\dbltextfloatsep}{8pt}

\begin{figure*}[t]
\centering
\includegraphics[width=0.98\textwidth]{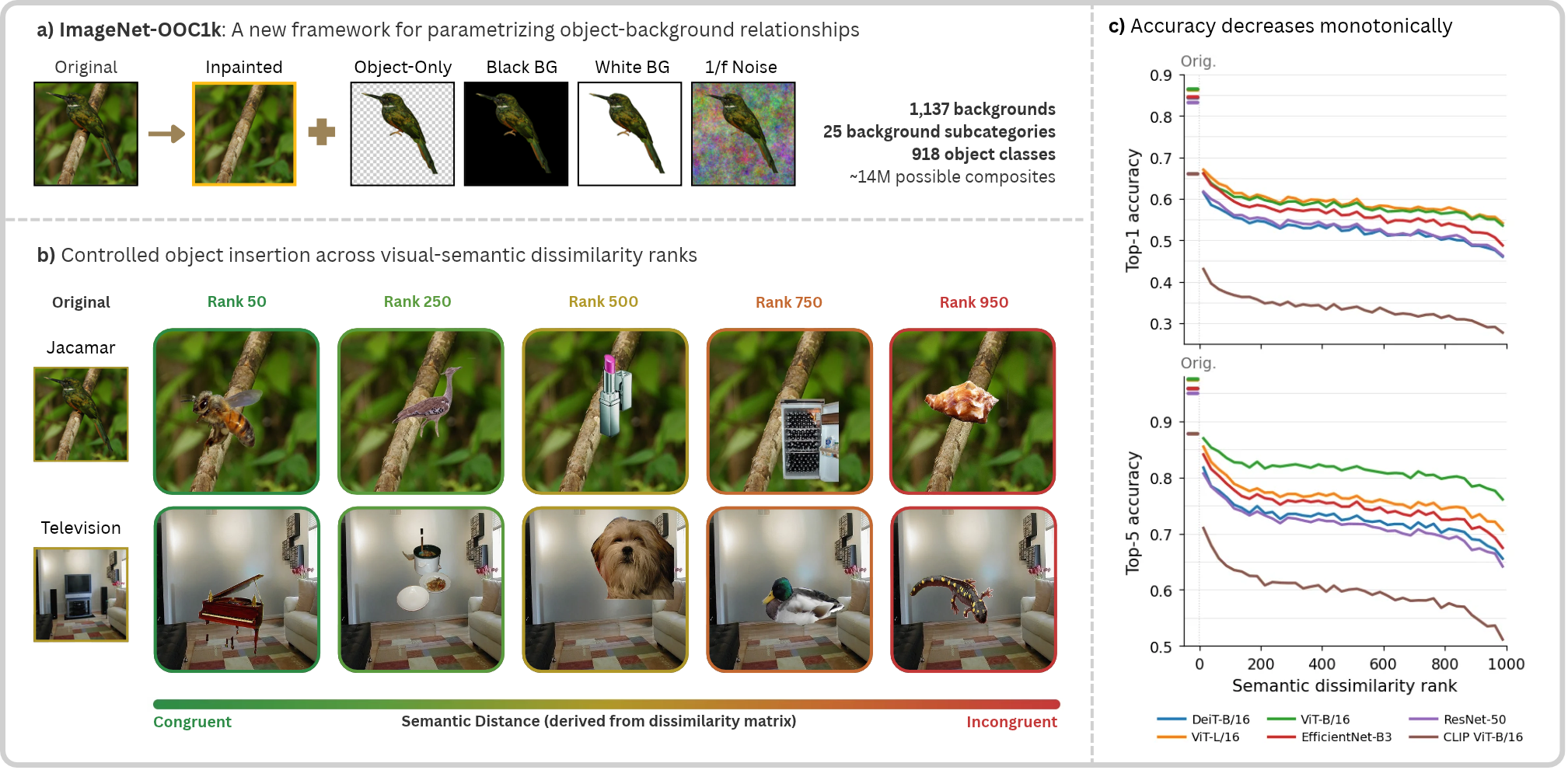}
\caption{\textbf{ImageNet-OOC1k enables controlled recomposition along a visual-semantic axis.}
\textbf{(a)}~Dataset overview. ImageNet-S masks~\cite{imagenetS} are used to construct original images, inpainted backgrounds, and object-only variants, including crops and composites on black, white, and $1/f$ noise backgrounds.
\textbf{(b)}~Objects are inserted at increasing semantic distance ranks relative to the original class. The example backgrounds (\emph{jacamar}, \emph{television}) use the same ranks (50, 250, 500, 750, 950), but different inserted objects because distance is defined relationally.
\textbf{(c)}~Top-1 accuracy decreases monotonically with semantic distance rank (bin width 25) for representative models (ViT-B/16, ViT-L/16, DeiT-B/16, ResNet-50, EfficientNet-B3, CLIP ViT-B/16). Colored markers (\emph{Orig.}) show original-image accuracy. Top-5 accuracy follows the same trend, indicating that semantic mismatch affects the broader class ranking.}
\label{fig:dataset}
\end{figure*}

\endgroup
\vspace{-6pt}

\section{Introduction}
Object recognition in natural images is often framed as an object-centric task \cite{imagenet, imagenet2}. In practice, however, scenes contain strong statistical regularities between objects and their surrounding context, and modern vision models exploit these regularities extensively \cite{zhu2017object, beery2018recognition, geirhos2020shortcut}. Existing benchmarks study this dependence using binary or coarse manipulations, such as contrasting congruent and incongruent object--background pairings \cite{sagawa2020investigation, xiao2021background, malik2024objectcompose}. While these approaches show that context matters, they do not provide a continuous notion of semantic relatedness. This raises the central question of what happens to recognition as object--background mismatch increases.

We address this limitation by introducing \textbf{ImageNet-OOC1k}\footnote{Code and dataset information: \url{https://github.com/ToastyDom/ImageNet-OOC1k}}, a benchmark for graded, parametrically controlled manipulation of object--background relationships. We define a visual-semantic axis over ImageNet classes from a consensus of classifier representations across multiple pretrained models, whose classifier heads can be interpreted as class prototypes in feature space that encode semantic structure \cite{vilas2023}. This allows recognition to be analyzed directly as a function of semantic distance.

We then use this classifier-derived axis to measure how recognition changes across 35 models. Across convolutional neural networks, Vision Transformers, and vision--language models, accuracy follows a monotonic function of semantic object--background distance (Figure~\ref{fig:dataset}c). The closely aligned trajectories indicate that context sensitivity follows a shared and structured pattern rather than arising from isolated failures.

To investigate the mechanisms underlying this behavior, we analyze internal representations in Vision Transformers, which allow patch-level inspection of class evidence. Using ViT-B/16 as a case study, we show that a subset of failures arise despite spatially distributed patch-level evidence for the correct object, consistent with incomplete consolidation of competing object and background signals into the final prediction. We examine their recoverability with a spatial re-ranking probe that leverages intermediate spatial structure, improving in-sample accuracy by +5.7 percentage points while maintaining a 9.4:1 found-but-lost (FBL) recovery-to-breakage ratio. This indicates that relevant information remains available beyond what is reflected in the final prediction.

Our contributions are threefold: (1) we introduce ImageNet-OOC1k, a benchmark for graded control over object--background relationships along a continuous visual-semantic axis; (2) we identify a highly consistent monotonic relationship between semantic distance and recognition accuracy across diverse vision models; and (3) we identify a failure mode under context shifts that suggests limitations in how Vision Transformers aggregate available object evidence, and demonstrate that a substantial fraction of these errors can be recovered through spatial re-ranking.

\section{Related Work}

We discuss prior work most related to ours, focusing on context dependence in visual recognition, datasets for studying object--background relationships, and methods for analyzing model internals.

\textbf{Object--background context in visual recognition.}
A large body of work shows that recognition models rely on contextual information beyond the object itself. Backgrounds alone can carry significant class signal \cite{moayeri2022comprehensive,zhu2017object}, and models often fail when objects appear in atypical environments \cite{beery2018recognition, rosenfeld2018elephant}. Controlled evaluations further show substantial performance drops when contextual cues are altered or removed \cite{barbu2019objectnet, xiao2021background}. These effects are often framed as shortcut learning, where models exploit spurious correlations in the data, including correlations between object classes and backgrounds \cite{geirhos2020shortcut, sagawa2020investigation}.

\textbf{Datasets for context sensitivity.}
Existing datasets study context through binary or coarse manipulations, such as contrasting congruent and incongruent pairings or introducing controlled correlations between objects and environments \cite{sagawa2020investigation, xiao2021background}. Other approaches explore unusual object--scene relationships or induce distribution shifts by placing objects in atypical environments \cite{kattakinda2022focus, beery2018recognition, rosenfeld2018elephant, janouskova2025robustcontextawareobjectrecognition}. Recent generative and inpainting-based pipelines increase contextual diversity \cite{malik2024objectcompose, yang2026coinco}, while synthetic datasets allow controlled semantic or physical violations \cite{Bomatter_2021_ICCV}. However, these approaches lack precise continuous control over semantic relatedness and therefore cannot characterize how recognition varies with the degree of mismatch.

\textbf{Mechanistic interpretability in Vision Transformers.}
Within the growing literature on vision model interpretability, our work builds most directly on logit lens methods that decode intermediate predictions and reveal how class evidence evolves across layers \cite{nostalgebraist2020logitlens, vilas2023, gandelsmaninterpreting}. Related work highlights limitations of the CLS token, showing that it can fail to capture object information in complex scenes \cite{khajuria2025}. However, these tools have not been applied to controlled context manipulations. We connect them to context sensitivity by testing whether failures under controlled object--background mismatch reflect limitations in aggregating distributed object evidence.

\section{ImageNet-OOC1k: Parametrizing Object--Background Relationships}
\label{sec:3_ImageNet_OOC}

Understanding how recognition changes under object--background mismatch requires precise control over the semantic relationship between objects and scenes. Existing benchmarks mainly use binary or coarse manipulations, which do not allow performance to be analyzed as a continuous function of mismatch. To address this, we introduce ImageNet-OOC1k, a benchmark that treats object--background context as a quantitative variable and enables recognition to be analyzed across the spectrum from compatible to strongly mismatched pairings (Figure~\ref{fig:dataset}a,b). Full dataset details are provided in Appendix~\ref{sec:appendix_ooc}.

\subsection{Dataset Construction}

ImageNet-OOC1k builds on ImageNet-S~\cite{imagenetS}, which provides high-quality segmentation masks for images from the ImageNet validation set \cite{imagenet, imagenet2}. Using these annotations, each image is decomposed into a foreground object and a background obtained via human-supervised inpainting. We use the \texttt{IOPaint} toolkit~\cite{iopaint2024} with the MAT model~\cite{li2022mat}, resulting in 1,137 background-only scenes.

The same masks are used to construct a reusable object library of foreground crops aligned with ImageNet class labels, spanning 918 classes, of which 915 contain at least four valid object instances. The 1,137 backgrounds are grouped into nature scenes (558 images, 12 subcategories), human-related scenes (533 images, 13 subcategories), and miscellaneous scenes (46 images) containing close-ups, ambiguous scenes, or backgrounds that do not fit either category. Example backgrounds are shown in Appendix Figure~\ref{fig:object_cats}, and detailed dataset statistics are provided in Appendix Table~\ref{tab:ooc_detailed}.

ImageNet-OOC1k contains inpainted backgrounds, object masks, composite specifications, and metadata describing object classes, scene categories, and generation parameters. Object-only control images are described in Appendix~\ref{sec:object-only-im_gen}. Because the images derive from ImageNet/ImageNet-S, image-derived components are released in forms compatible with source-data terms, as detailed in Appendix~\ref{sec:appendix_release}.

\subsection{Controlled Composite Generation}

The key property of ImageNet-OOC1k is the explicit decoupling of objects and backgrounds, enabling combinatorial recomposition of scenes. For each background, object classes are combined across the full range of relative positions along the visual-semantic axis (Section~\ref{sec:consensus}), spanning a range from highly compatible to strongly mismatched pairings.

This defines approximately 14 million possible object--background combinations using individual object instances, enabling dense sampling across visual-semantic space rather than relying on a small set of predefined conditions. Object sampling, spatial placement, and scale follow fixed, reproducible procedures, and all generation parameters are recorded. Composite-generation details, including sampling, scaling, and placement, are provided in Appendix~\ref{sec:appendix_comp_gen_det}. The automated generation pipeline is available at \url{https://github.com/ToastyDom/ImageNet-OOC1k}, together with metadata, composite specifications, and documentation of the source-data constraints. This treats object--background context as a graded experimental variable, allowing recognition to be measured as a function of visual-semantic distance rather than under fixed congruent or incongruent conditions (Figure~\ref{fig:dataset}b).

\begin{figure}[t]
    \centering
    \includegraphics[width=\textwidth]{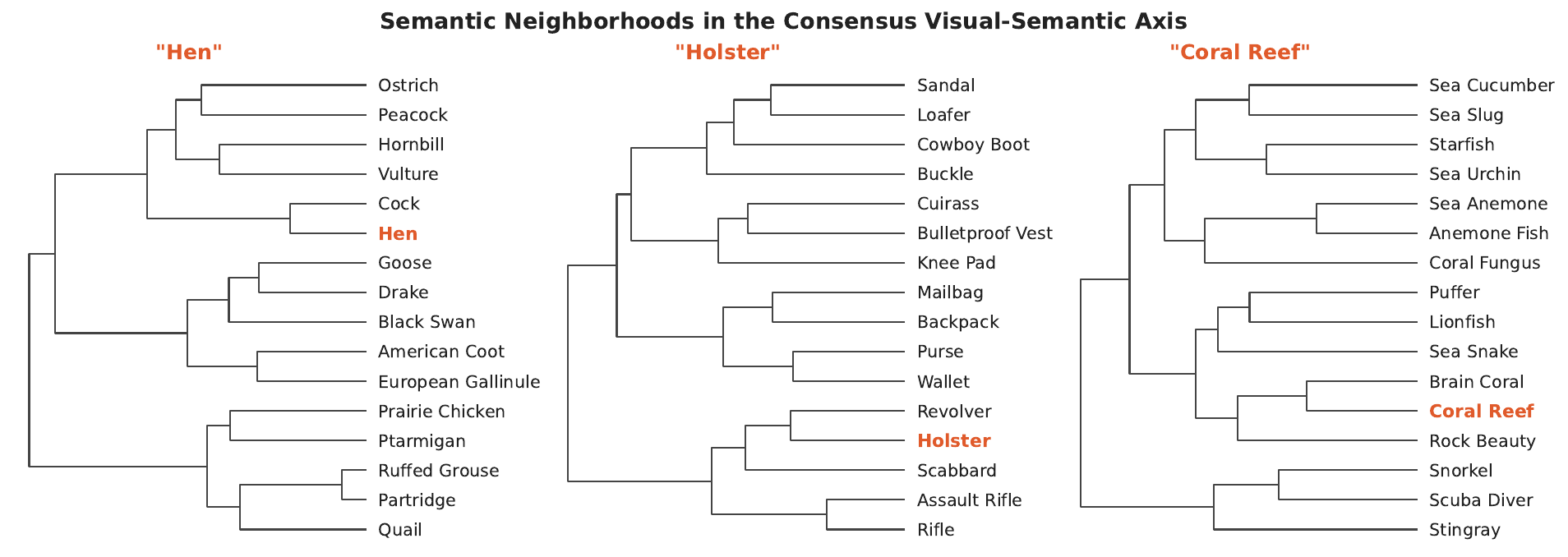}
    \vspace{-8pt}
    \caption{
    \textbf{Semantic neighborhoods defined by the consensus visual-semantic axis.}
    Each panel shows the 15 nearest neighbors of a focal class (highlighted), clustered using Ward linkage.
    Classes form coherent semantic groups (e.g., \textit{hen} with birds, \textit{holster} with weapons, \textit{coral reef} with marine organisms), illustrating that classifier representations encode rich semantic structure.
    Similar high-level groupings are consistently observed across individual classifiers (Appendix Figures~\ref{fig:supp_vit}--\ref{fig:supp_effnet}), indicating that this structure is broadly shared across models.
    }
    \label{fig:dendrograms}
\end{figure}

\section{A Consensus Visual-Semantic Axis}
\label{sec:consensus}

To parameterize object--background relationships in ImageNet-OOC1k, we define semantic distance directly in the representational space of pretrained classifiers. The final linear layer of modern models can be interpreted as a set of class prototypes, where each row corresponds to a class embedding in feature space \cite{vilas2023}. The relative arrangement of these prototypes forms a structured embedding in which visually and semantically related classes lie nearby. This organization emerges from visual discrimination and is consistently observed across architectures, as shown in Figure~\ref{fig:dendrograms} and Appendix~\ref{sec:appendix_semantic_structure} (Figures~\ref{fig:supp_consensus}--\ref{fig:supp_effnet}).

To obtain a stable and model-agnostic estimate, we aggregate across diverse architectures. We compute semantic distances from ten pretrained ImageNet classifiers, including Vision Transformers (ViT-B/16 at 224 and 384 resolution, ViT-S/16, ViT-L/16, ViT-B/32) and convolutional networks (ResNet-50, ResNet-101, EfficientNet-B3, EfficientNet-B5, DenseNet-121). For each model \( m \), let \( U^{(m)} \in \mathbb{R}^{C \times d} \) denote the classifier weight matrix over \(C\) classes and \(d\)-dimensional class embeddings, with row \(u_i^{(m)}\) representing class \(i\). Rows are L2-normalized and pairwise cosine dissimilarity is computed,
\begin{equation}
D_{ij}^{(m)} = 1 - \frac{u_i^{(m)}}{\|u_i^{(m)}\|_2} \cdot \frac{u_j^{(m)}}{\|u_j^{(m)}\|_2}.
\end{equation}
The resulting matrices are symmetrized via \( (D + D^\top)/2 \) to correct minor numerical asymmetries. To account for scale differences across the \(M\) models, we z-score the off-diagonal entries of each symmetrized matrix to obtain \(\tilde{D}^{(m)}\). Averaging the normalized matrices gives the consensus dissimilarity matrix,
\begin{equation}
D_{ij}^{\text{cons}} = \frac{1}{M} \sum_{m=1}^{M} \tilde{D}_{ij}^{(m)}.
\end{equation}
The consensus matrix is then symmetrized, and its off-diagonal entries are min--max rescaled to \( [0,1] \). Diagonal entries are fixed at zero throughout, preserving \(D_{ii}^{\text{cons}} = 0\). The resulting consensus axis summarizes this shared class structure while reducing model-specific noise. Low values correspond to related classes within this discriminative space, while high values reflect increasing mismatch. This axis defines the continuous variable used to parameterize object--background relationships in ImageNet-OOC1k. We assess potential circularity through model-overlap and own-axis controls and comparisons with axes derived from independently pretrained representations (Appendix~\ref{sec:appendix_distance_validation}). The decline generalizes across the tested discriminative axes, while retrieval, linguistic, and taxonomic distances produce different curves.

\begin{figure}[t]
    \centering
    \includegraphics[width=\textwidth]{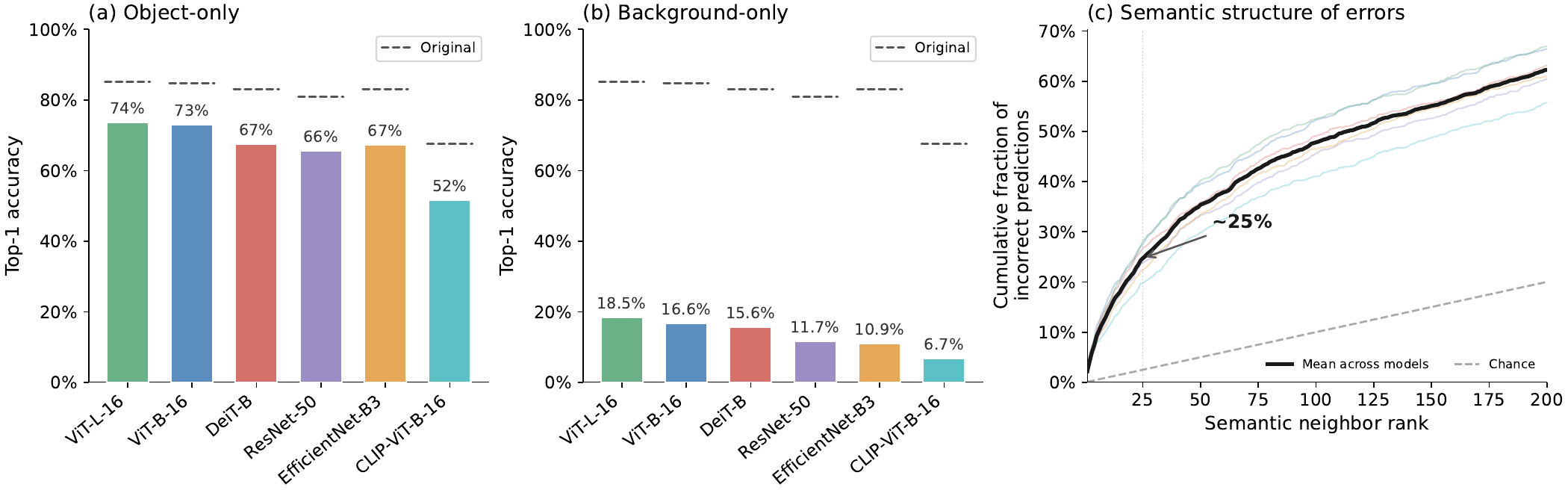}
    \vspace{-12pt}
    \caption{
    \textbf{Objects and backgrounds provide separable recognition signals.}
    \textbf{(a)}~Top-1 accuracy for object-only images across representative models. Bars show performance averaged across three neutral background conditions (black, white, and $1/f$ noise; see Appendix~\ref{sec:object-only-im_gen}). Dashed lines indicate accuracy on original images.
    \textbf{(b)}~Top-1 accuracy for background-only images across the same models. Dashed lines again indicate original-image accuracy.
    \textbf{(c)}~Cumulative semantic rank of incorrect background-only predictions. The x-axis shows semantic neighbor rank relative to the ground-truth class, where 1 denotes the nearest semantic neighbor. The y-axis shows the cumulative fraction of incorrect predictions whose predicted class falls within that rank. Thin lines correspond to the models shown here, and the thick line shows their mean. The dashed line indicates chance-level accumulation. Results for all 35 models are shown in Appendix~\ref{sec:AppC1_Object_BG_Signal}.
    }
    \label{fig:obj_bg_semantics}
\end{figure}

\section{Behavioral Analysis of Context Dependence}
\label{sec:5_Behavioral_Analysis}

Having established a continuous visual-semantic axis over object--background relationships, we now analyze how context influences recognition. We ask three questions. First, do backgrounds alone induce structured, class-specific predictions? Second, how does recognition change as a function of semantic mismatch? Third, do these effects depend on background type?

\subsection{Backgrounds Induce Structured Semantic Predictions}
\label{sec:results_context_signal}

We begin with a control analysis that separates object-derived from background-derived evidence. Prior work has shown that backgrounds can carry class signal \cite{zhu2017object,moayeri2022comprehensive}. Here, we quantify this effect within the same semantic space used throughout the paper, using object-only images composited onto neutral backgrounds (black, white, and $1/f$ noise), with an example in Figure~\ref{fig:dataset}a and details in Appendix~\ref{sec:object-only-im_gen}.

Removing natural context reduces object-only performance by approximately 12--16 percentage points across models (Figure~\ref{fig:obj_bg_semantics}a). This indicates that models rely primarily on object information, but still benefit from contextual support. Performance varies little across the three neutral background conditions, is consistent across all 35 models (Table~\ref{tab:model_list}; Appendix~\ref{sec:AppC1_Object_BG_Signal}, Figure~\ref{fig:obj_all_comb}), and shows similar trends under top-5 evaluation. In contrast, background-only images yield substantially lower but still surprisingly high top-1 accuracy (Figure~\ref{fig:obj_bg_semantics}b), reaching 13.3\% on average across the six representative models despite the complete absence of the object. Under top-5 evaluation, background-only accuracy reaches up to $\sim$40\% across the full 35-model set, with the highest values among BEiT and ConvNeXt models and lower values for CLIP and the evaluated CNNs (Appendix~\ref{sec:AppC1_Object_BG_Signal}).

Crucially, these predictions are not arbitrary. Among background-only images with an incorrect top-1 prediction, approximately 25\% of predicted classes fall within the 25 nearest visual-semantic neighbors of the ground-truth class (Figure~\ref{fig:obj_bg_semantics}c). Across the full 35-model set, the corresponding analysis yields an 8.9-fold enrichment over a 2.9\% permutation baseline ($p < 0.001$, Appendix~\ref{sec:AppC1_Object_BG_Signal}). Thus, background signals reliably place the model in the correct semantic neighborhood, even when the exact class is not recovered.

Together, these results show that backgrounds provide structured, class-relevant information rather than noise. We next ask how recognition changes when object- and background-derived signals are systematically pulled apart along the visual-semantic axis.

\subsection{Recognition Degrades with Semantic Distance}
\label{Section:Semanticdistance}

As object and background semantics are pulled apart, accuracy decreases monotonically with semantic object--background distance across representative models (Figure~\ref{fig:dataset}c). The gap to original-image accuracy should not be interpreted as part of this decline across composite bins, since the lowest-rank bin already averages composites from the 25 nearest semantic neighbors and may include recomposition costs from scaling, insertion, and residual artifacts. This pattern generalizes across all 35 models, despite differences in architecture, training objective, and baseline performance (Appendix~\ref{sec:appendix_semantic_distance}). Across controlled composites (Appendix~\ref{sec:appendix_comp_gen_det}), top-1 accuracy drops by approximately 9--19 percentage points from the most similar to the most dissimilar pairs. Image-level logistic regressions confirm that this trend is not an artifact of binned visualization, with strictly negative slopes and 95\% confidence intervals below zero for every model (Appendix~\ref{sec:appendix_semantic_distance}). Negative slopes persist across all 35 models after controlling for placement, scale, and background and object-instance effects, and after additionally controlling for boundary discontinuities on a separate 30,000-composite sample (Appendix~\ref{sec:compositing_artifacts}). The trend also extends to top-5 accuracy, showing that semantic mismatch reshapes the broader class ranking rather than only the top prediction. CLIP~\cite{clip_radford} follows the same trend despite multimodal training, indicating that language grounding does not eliminate reliance on visual context.

While the overall relationship is monotonic, backgrounds separate into two regimes. Most induce decreases in accuracy, while a smaller subset shows facilitation, where accuracy increases with semantic distance. These regimes are structured, with nature scenes primarily driving interference and human-related environments more often producing facilitation (Appendix~\ref{sec:appendix_het_across_bg}). This may reflect the greater visual consistency of nature scenes, which impose stronger contextual expectations than more heterogeneous human-related environments. A mechanistic analysis of these background-specific regimes is beyond the scope of this paper, but the pattern shows that context type modulates object--background interactions in distinct ways.

Finally, this monotonic decline depends on the visual-semantic space used to define distance. Classifier heads and linear probes on independently pretrained DINOv1 \cite{caron2021emerging}, DINOv2 \cite{oquab2024dinov2learningrobustvisual}, and MAE features \cite{he2022masked} reproduce the decline, whereas retrieval, CLIP-text, and WordNet \cite{wordnet} distances produce qualitatively different curves (Appendix~\ref{sec:appendix_distance_validation}). Thus, the result is not a generic consequence of ranking classes by any semantic measure, but reflects discriminative visual class geometry.


\section{Representational Competition and Aggregation Failure}
\label{sec:mechanistic}

Section~\ref{Section:Semanticdistance} showed that recognition accuracy declines monotonically as semantic object--background distance increases, suggesting that failures reflect structured interactions between object and background signals. We therefore ask whether these failures arise because object information is missing, or because competing object- and background-derived signals are not successfully combined.

We study this question in ViT-B/16, where the CLS token serves as a global aggregation token that collects information from spatial patch tokens for the final classification decision. This architecture enables patch-level inspection via the logit lens~\cite{nostalgebraist2020logitlens, vilas2023}, allowing us to track how object- and background-derived class evidence evolves across layers before it is consolidated into the final prediction. The analysis is performed on a stratified subset of 75,000 composites sampled uniformly across 20 semantic distance bins (Appendix~\ref{sec:appendix_comp_gen_det}).

\begin{figure*}[t]
\centering
\includegraphics[width=\textwidth]{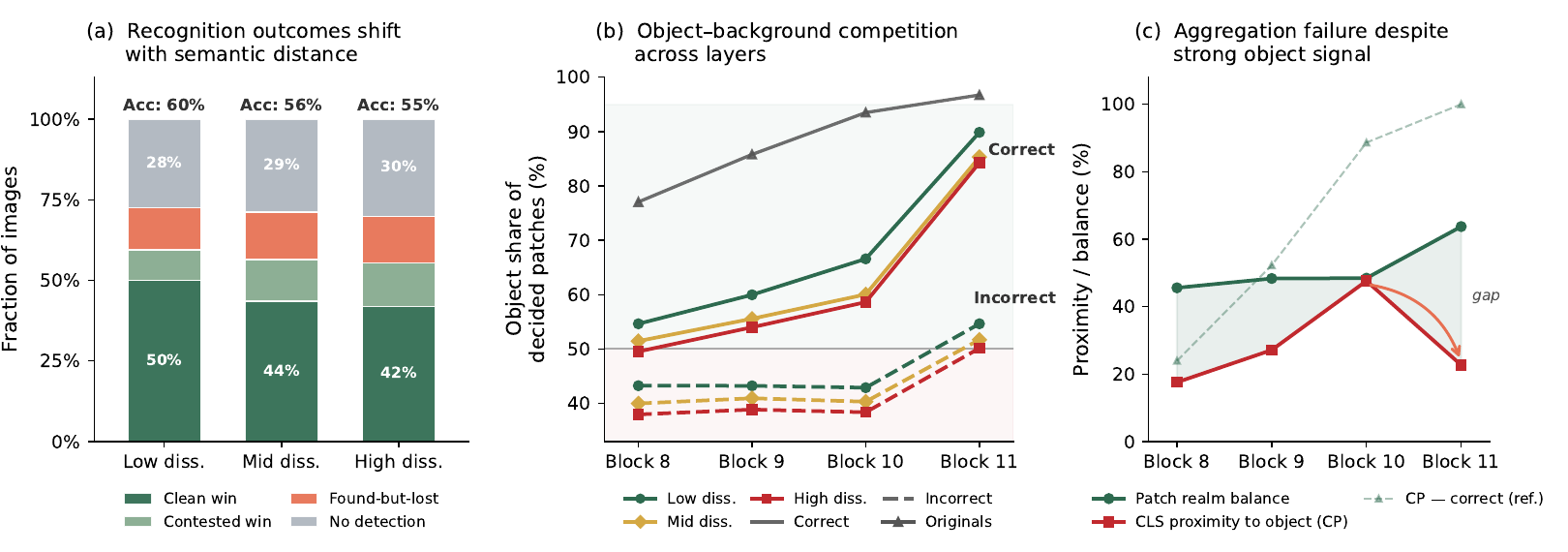}
\vspace{-8pt}
\caption{
\textbf{Recognition outcomes reveal increasing object--background competition and aggregation failure.}
All panels use the 75,000-composite stratified subset analyzed for ViT-B/16.
\textbf{(a)}~Outcome composition across all composites, grouped into low (1--350), mid (351--700), and high (701--1,000) semantic dissimilarity rank bands. Stacked bars show the fraction of images assigned to each outcome category at post-norm.
\textbf{(b)}~Object share of decided patches (realm balance) across blocks 8--11, computed over all composites and split by correctness and dissimilarity group. Realm balance captures the competition between object- and background-aligned patch evidence. Correct predictions transition to object dominance at block 11, whereas incorrect predictions remain balanced or background-leaning.
\textbf{(c)}~Aggregation dynamics restricted to found-but-lost cases. The green line shows patch-level realm balance, and the red line shows CLS classification proximity (CP) to the object class. Patch evidence increasingly favors the object, while CLS proximity to the object drops at the final layer.
}
\label{fig:mechanistic}
\end{figure*}

\subsection{Semantic Distance Increases Object--Background Competition}

We decompose predictions into four outcome types based on correctness and patch-level object evidence at the final layer. Clean wins are correct predictions with strong object dominance across patches. Contested wins are correct predictions with weaker or mixed patch evidence. Found-but-lost cases are incorrect predictions in which many patches nevertheless predict the object class. No-detection cases are incorrect predictions with little patch-level object evidence. Formal definitions of the outcome taxonomy are given in Appendix~\ref{sec:appendix_taxonomy}. The metrics used to define them, including semantic pull, realm balance, and classification proximity, are described in Appendix~\ref{Coremetrics}.

We first ask how the frequency of these outcome types changes with semantic distance. Figure~\ref{fig:mechanistic}a shows that clean wins decrease as object--background distance increases, while no-detection cases account for most of the performance drop. Contested wins and found-but-lost (FBL) cases change more modestly in frequency, but frequency alone does not capture how the underlying evidence changes. We therefore quantify object--background competition using \textit{realm balance}, the fraction of decided patches favoring the object rather than the background. Figure~\ref{fig:mechanistic}b shows that increasing semantic distance shifts predictions toward background dominance, especially for incorrect cases, which remain background-leaning until the final layer. At the final layer, correct predictions transition sharply to object dominance, whereas incorrect predictions remain unresolved. Successful recognition is therefore associated with object evidence overcoming sustained background competition at the aggregation stage.

\subsection{Found-but-Lost Cases Reveal Aggregation Failure}
\label{sec:agg_failure}

\begin{figure}[t]
\centering
\includegraphics[width=0.95\textwidth]{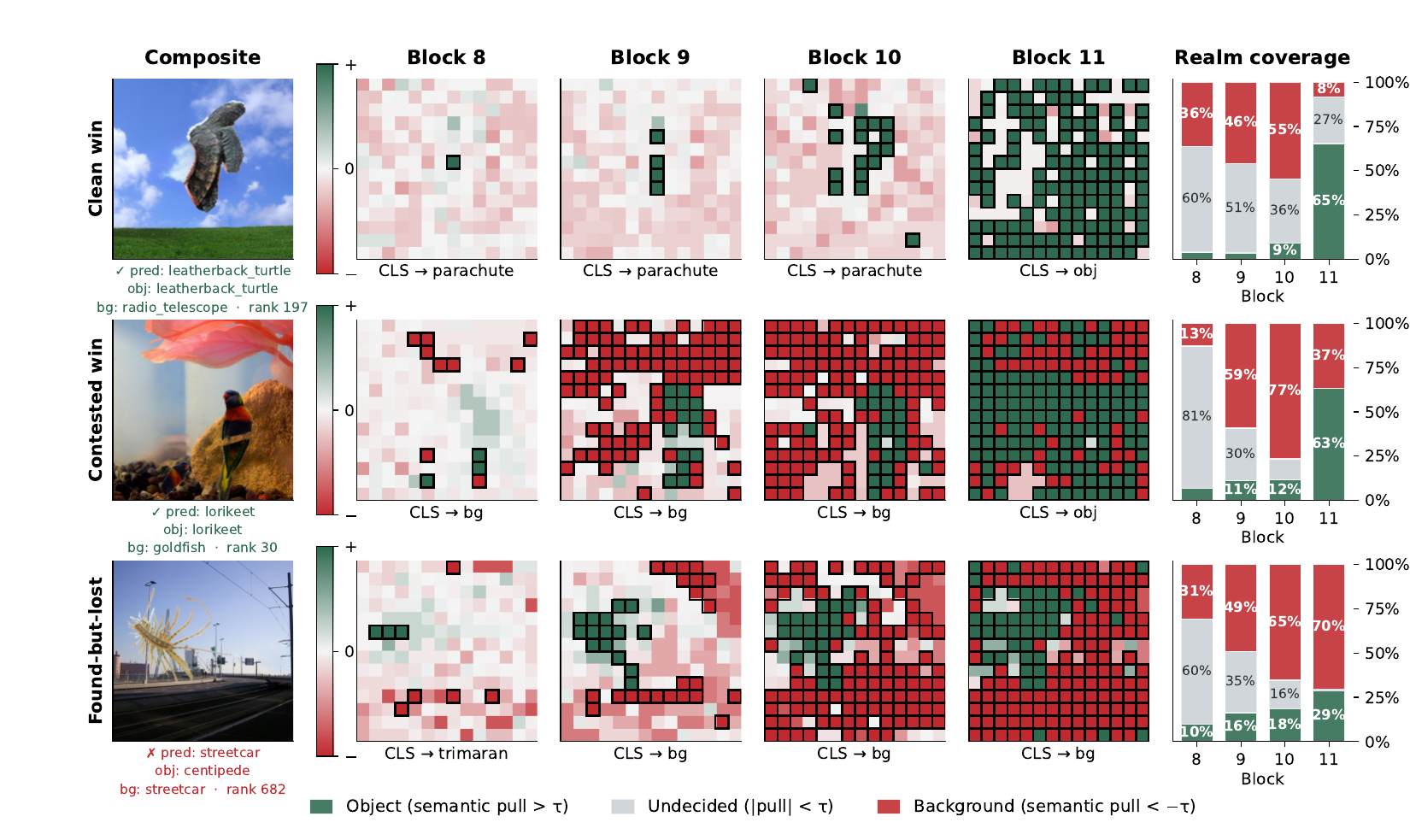}
\vspace{-6pt}
\caption{
\textbf{Patch-level semantic pull heatmaps across outcomes.}
Each row shows an example composite and patch-level predictions across the final transformer blocks (8--11). Green indicates alignment with the object class and red with the background class, with intensity reflecting semantic proximity. Black borders denote exact class matches. The rightmost column shows the fraction of patches assigned to each region. Rows correspond to clean win, contested win, and found-but-lost cases, illustrating full, partial, and stalled expansion of object-aligned patches.
}
\label{fig:heatmaps}
\end{figure}

FBL cases represent one specific error type in which the object is clearly detected at the patch level, yet the model still predicts the wrong class. They provide a targeted diagnostic for aggregation failure, while the causal mechanism remains unresolved. The spatial structure of these cases shows that the key difference is not whether the object is found, but how object evidence evolves across the patch grid (Figure~\ref{fig:heatmaps}). In clean wins, object evidence spreads across most of the grid by the final layer and achieves global dominance. In contested wins, this expansion is partial. In FBL cases, expansion stalls at an intermediate stage, leaving object evidence spatially localized, while no-detection cases show little to no expansion. Appendix Figure~\ref{fig:appendix_FBL_fix}b quantifies this pattern by measuring the expansion of exact object-class patches across layers.

Figure~\ref{fig:mechanistic}c shows how this incomplete expansion relates to the CLS representation. Up to block 10, patch-level evidence and the CLS representation evolve in parallel toward the object class. At the final layer, however, this alignment breaks. Patch-level evidence continues to strengthen toward the object, while the CLS representation reverses direction, with classification proximity (Appendix~\ref{Coremetrics}) dropping from approximately 48\% to 23\%. Thus, FBL cases reveal a failure to consolidate available object evidence into the final prediction.

This failure is quantitatively a near-miss. In 98.7\% of FBL cases, the correct class appears within the CLS top-5 at post-norm with a median rank of 2, compared to 37\% and a median rank of 9 for no-detection cases (Appendix Figure~\ref{fig:appendix_FBL_fix}a). This shows that the correct class is often available in the model’s representation, but not selected.

The incomplete expansion also leaves a measurable spatial signature. In FBL cases, patches supporting the runner-up class, often the true object class, form a compact cluster, while patches supporting the predicted class are more dispersed, especially at block 10 where aggregation remains unresolved (Appendix Figure~\ref{fig:appendix_FBL_fix}c). While not all errors are of this type, these results identify a clear and interpretable failure mode characterized by incomplete expansion and consolidation of available object evidence into the final decision.

\subsection{Recoverability of Aggregation Failure}

The compact spatial signature identified above suggests that some aggregation failures may be recoverable from intermediate representations. We exploit this structure as a diagnostic probe, rather than as a deployable inference-time correction method. FBL cases exhibit three properties that make this probe possible. The correct class often remains near the top of the CLS ranking, object evidence remains spatially compact at intermediate layers, and aggregation is uncertain near the final decision. These observations motivate a constrained re-ranking strategy that compares the predicted class with the runner-up class using logit uncertainty, spatial compactness, and semantic distance.

\begin{figure}[t]
\centering
\includegraphics[width=0.9\linewidth]{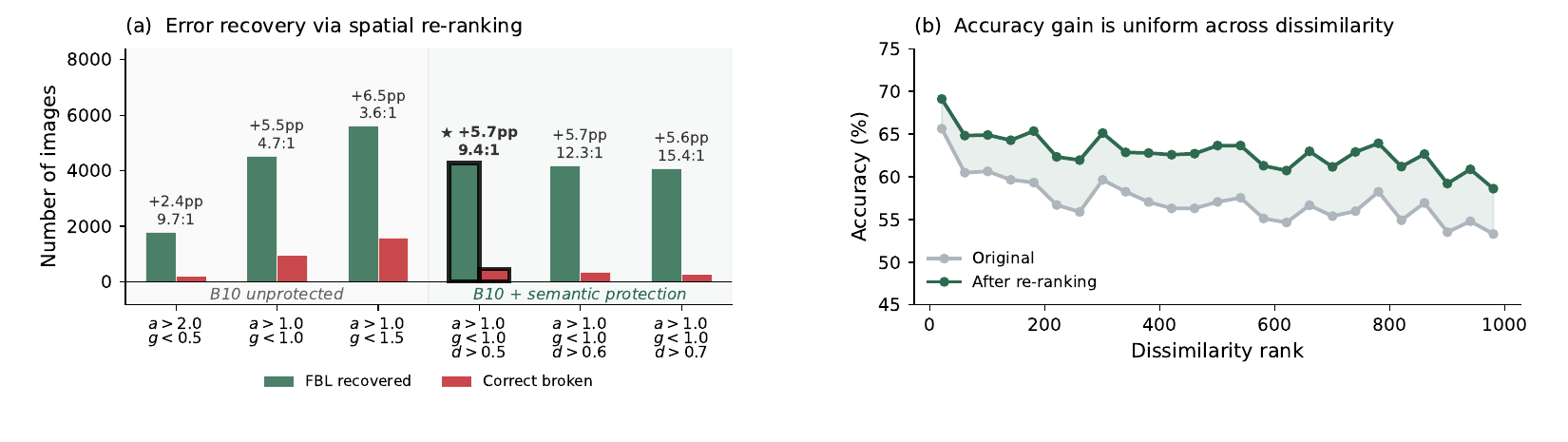}
\vspace{-6pt}
\caption{
\textbf{Spatial re-ranking probes the recoverability of aggregation failures in ViT-B/16.}
\textbf{(a)}~Recovery and breakage counts at block 10 across re-ranking rules, measured as FBL cases recovered and originally correct predictions broken. The left panel uses spatial compactness advantage and logit gap ($a$, $g$); the right additionally constrains semantic distance between the top-1 and runner-up classes ($d$) to reduce near-neighbor swaps. The star marks the selected operating point used in the main analysis.
\textbf{(b)}~Top-1 accuracy before and after applying the selected re-ranking rule across semantic distance ranks (bin width 40), evaluated on the 75,000-image stratified ViT-B/16 diagnostic subset, which explains the slight baseline difference from the full composite benchmark in Figure~\ref{fig:dataset}c. Re-ranking improves accuracy across the full range, with a net gain of approximately +5.7 percentage points.
}
\label{fig:fbl_recovery}
\end{figure}

The re-ranking rule combines CLS uncertainty, measured by the logit gap $g$, spatial compactness advantage between the top-1 and runner-up predictions $a$, and semantic distance between the competing classes $d$ (Appendix~\ref{app:reranking}). Across the threshold sweep in Figure~\ref{fig:fbl_recovery}a, adding the semantic-distance constraint improves the recovery-to-breakage tradeoff, and the selected operating point recovers 4,277 FBL cases while breaking 456 correct predictions. This yields a 9.4:1 FBL recovery-to-breakage ratio. Including incidental corrections of non-FBL errors, the same rule produces an in-sample net accuracy improvement of +5.7 percentage points on the ViT-B/16 diagnostic subset, with gains observed across the full semantic distance range (Figure~\ref{fig:fbl_recovery}b).

Each component reflects the aggregation-failure signature identified above. The logit gap identifies uncertain aggregation, spatial compactness captures localized object evidence, and semantic distance prevents near-neighbor swaps. To test whether this signature is specific to ViT-B/16 or to CLS-token pooling, we repeated the outcome taxonomy, patch-expansion analysis, and fixed-rule recovery probe across ViT-S/16, ViT-L/16, DeiT-B/16, and ViT-RelPos-B/16, a no-CLS ViT with average pooling. Across models, FBL cases remained near-miss errors, with the correct class appearing in the top-5 in 94.7--98.7\% of cases. The fixed rule produced positive net accuracy changes across architectures, including in the average-pooling model, with recovery magnitude varying with the strength of the spatial compactness signal (Appendix~\ref{app:tax_gen}--\ref{app:fix_gen}). Together, these results are consistent with the interpretation that a substantial subset of failures reflects how information is combined, rather than missing object representations.

\section{Discussion and Conclusion}
\label{sec:7_Discussion}

We introduced ImageNet-OOC1k, a framework for graded, model-grounded control over object--background relationships. This setting reveals that recognition accuracy decreases monotonically with visual-semantic object--background distance across diverse vision models, showing that context sensitivity is a predictable property rather than a discrete failure under incongruence.

The results also show that backgrounds provide structured, class-relevant signals that actively shape predictions. Recognition can therefore be understood as a competition between object-derived and background-derived evidence, whose balance changes systematically with visual-semantic distance. Individual backgrounds nevertheless differ, with some showing facilitation rather than interference (Appendix~C.3). This graded structure is central to the benchmark. By separating object--background relationships along this axis, ImageNet-OOC1k exposes failure signatures that binary congruent/incongruent manipulations would likely conflate. In Vision Transformers, we identify one such signature, where a subset of failures points to limitations in how competing signals are aggregated. Even when the correct class is strongly represented across patches, the final prediction can fail to consolidate this evidence. The recovery experiment supports this interpretation, showing that a substantial fraction of FBL errors can be corrected using intermediate spatial structure. While this probe is not a complete solution, it suggests that aggregation is a potential bottleneck and points to future architectural mechanisms for better resolving object--background competition.

Our work has several limitations. The dataset relies on inpainting and compositing, which may introduce artifacts, and object cutouts inherited from ImageNet-S are not always perfectly coherent. Placement, scale, and appearance changes are sampled independently of the visual-semantic axis. The decline persists after controlling for placement, scale, and background and object-instance effects, and after additionally controlling for boundary discontinuities on a separate sample (Appendix~\ref{sec:compositing_artifacts}). Pair-specific visual compatibility between object and scene, however, cannot be fully separated from semantic mismatch. The single-target setting also leaves multi-object interactions for future investigation. Why some backgrounds hinder recognition while others help remains unclear. Differences in object plausibility and scene diversity may play a role. While the behavioral result spans all 35 models, the mechanistic evidence is restricted to the tested Vision Transformers. Generalization to CNNs remains an open question requiring an analogous spatial diagnostic. The axes remain tied to ImageNet categories, including through the labels used to train the independent probes, and their alignment with human similarity judgments has not been tested. Finally, the recovery analysis requires access to intermediate representations and is intended as a diagnostic probe rather than a deployable inference-time method. It does not establish the causal mechanism of FBL errors.

Overall, ImageNet-OOC1k shows that controlled context manipulations can link recognition errors to patterns of object and background evidence within models. By turning object--background context into a graded experimental variable, the dataset can support future work on anticipating failures, comparing how architectures resolve competing evidence, and designing models that more reliably integrate objects within the scenes in which they appear.

\section*{Acknowledgments}
This work was partly funded by the BMFTR neuroAI grant GROWUPAI, DFG project 5368 (Abstract REpresentations in Neural Architectures, ARENA), and DFG project, RO 6458/5-1 (Learning from the Environment Through the Eyes of Children, LEECHI). We gratefully acknowledge support from the Hessian Center for Artificial Intelligence and Goethe University (NHR Center NHR@SW) for providing the computing and data-processing resources needed for this work. The authors declare no competing interests.

\bibliographystyle{plainnat}  
\bibliography{references}

\input{appendix.tex}





\end{document}

%% file: appendix.tex
\clearpage
\appendix

\clearpage
\section{ImageNet-OOC1k Dataset Details}
\label{sec:appendix_ooc}

ImageNet-OOC1k is derived from ImageNet-S \cite{imagenetS}, which provides segmentation masks for ImageNet validation images \cite{imagenet,imagenet2}. We use these masks to separate foreground objects from their original backgrounds. Inpainting was performed under human supervision using \texttt{IOPaint}~\cite{iopaint2024} with the MAT model~\cite{li2022mat}, producing background-only scenes while retaining the surrounding scene structure. For each image, we interactively selected the region to remove, repeated the inpainting when necessary, and retained only backgrounds that passed visual inspection. We adopted this procedure because automated mask-based inpainting often produced duplicated textures, implausible structures, or residual object content. We also provide a reusable library of segmented foreground objects. The resulting dataset contains 1,137 background-only scenes and an object library spanning 918 ImageNet classes.

This appendix provides the dataset statistics, composite-generation details, controls for compositing artifacts, object-only control construction, release information, and compute resources supporting the main experiments.

\subsection{Composite Generation and Experimental Subsets}
\label{sec:appendix_comp_gen_det}

The main composite experiments use the decoupled structure of ImageNet-OOC1k to sample a finite set of object--background pairings along the real-valued visual-semantic axis, with distance defined relative to each background's original class. Table~\ref{tab:ooc_detailed} summarizes the dataset composition, including background statistics, object library coverage, and the subsets used in the behavioral and mechanistic experiments. The full pool contains 1,137 backgrounds and 12,321 individual object instances, yielding 14,008,977 possible object--background pairings. This count refers to individual object instances rather than unique classes and does not include additional variations in placement or scale.

For the behavioral experiments (§\ref{Section:Semanticdistance}), we construct a large composite set from a stratified subset of 260 backgrounds, sampled as 10 backgrounds from each of 26 background groups: 12 nature subcategories, 13 human-related subcategories, and one miscellaneous group containing close-ups, ambiguous scenes, and backgrounds that do not fit either category.

For each object class, up to four instances are sampled from the object library when available. The object library spans 918 ImageNet classes. Classes with fewer than four valid instances are still included but contribute fewer composites. Objects are initially scaled by setting their dominant dimension, width for wide objects and height for tall objects, to a uniformly sampled 50--60\% of the corresponding background dimension while preserving aspect ratio. Each side is then clamped independently to a minimum of 100 pixels and a maximum of half the corresponding background dimension. Objects are placed at random positions with a 5\% margin from image borders.

Object brightness is scaled by a factor sampled uniformly from $[0.9,1.1]$ and colour saturation by a factor sampled uniformly from $[0.95,1.05]$. With probability $0.3$, objects are rotated by an angle sampled uniformly from $[-10^\circ,10^\circ]$. Placement, sampled scale, and appearance transformations are generated without reference to visual-semantic rank, and generation settings and per-composite metadata are recorded for reproducibility. This results in 953,862 composites covering the full visual-semantic distance range. Controls for potential compositing artifacts are described in Appendix~\ref{sec:compositing_artifacts}.

For the mechanistic analysis (§\ref{sec:mechanistic}), we use a stratified subset of 75,000 composites sampled uniformly across visual-semantic distance and background category (20 bins $\times$ 3 categories). Sampling is performed without balancing for correctness, preserving the natural distribution of model behavior within each condition.

The large underlying composite pool (Table~\ref{tab:ooc_detailed}) enables controlled sampling across visual-semantic space while maintaining diversity in object identity and scene structure.

\subsection{Robustness to compositing artifacts}
\label{sec:compositing_artifacts}

Placement and scale are sampled independently of visual-semantic rank, and each background spans nearly the full rank range. Nevertheless, realized object size, cutout properties, and compositing artifacts could still vary systematically with distance. We therefore test whether these properties explain the accuracy decline using the release dataset of 953,862 composites and all 35 models.

We first collect 18 generation properties from the composite metadata, generation code, and object masks, covering placement, scale, size constraints, and cutout geometry. Eccentricity measures distance from the image centre, while border distance measures the smallest gap between the bounding box and an image edge. Mask area is the fraction of the background covered by object pixels, and mask fill is the fraction of the cutout's bounding box occupied by those pixels. Outline complexity is the mask perimeter divided by the square root of its area. Aspect distortion measures the absolute log change in aspect ratio introduced by the size constraints.

For each property, we correlate its values across composites with their semantic ranks. This tests whether, for example, objects become systematically smaller or more peripheral as semantic distance increases. We also compare the property's mean between the first and last distance bins. To estimate the associated accuracy change, we multiply this difference in means by the univariate least-squares slope of accuracy on that property, using accuracy averaged across models. Table~\ref{tab:generation_properties} reports these measurements. Correlations near zero and similar first- and last-bin means indicate little systematic variation with distance, while the estimated contributions show the magnitude of the associated accuracy changes relative to the observed decline.

\begin{table}[!h]
    \centering
    \small
    \setlength{\tabcolsep}{5pt}
    \caption{
    \textbf{Generation properties across semantic distance.}
    First and last bins span ranks 1--25 and 976--1000. Contributions are univariate estimates in percentage points, with shares calculated relative to the observed change of $-14.46$ percentage points. Negative contributions follow the decline, while positive contributions oppose it and therefore have negative shares.
    }
    \label{tab:generation_properties}
    \begin{tabular}{@{}lrrrrr@{}}
        \toprule
        Property & $\rho$ &
        \shortstack{First-bin\\mean} &
        \shortstack{Last-bin\\mean} &
        \shortstack{Contribution\\(pp)} &
        \shortstack{Share\\(\%)} \\
        \midrule
        \multicolumn{6}{@{}l}{\textit{Placement}} \\
        Horizontal centre       & $\approx 0$ & 0.502 & 0.500 & $+0.001$ & $-0.01$ \\
        Vertical centre         & $\approx 0$ & 0.498 & 0.497 & $-0.005$ & $+0.03$ \\
        Eccentricity            & $+0.002$    & 0.173 & 0.174 & $-0.063$ & $+0.44$ \\
        Border distance         & $-0.001$    & 0.132 & 0.132 & $+0.001$ & $-0.01$ \\
        Rotation flag           & $-0.001$    & 0.301 & 0.296 & $+0.009$ & $-0.06$ \\
        \midrule
        \multicolumn{6}{@{}l}{\textit{Scale and size constraints}} \\
        Sampled scale factor      & $-0.001$ & 0.550 & 0.550 & $-0.005$ & $+0.03$ \\
        Relative width            & $-0.013$ & 0.457 & 0.448 & $-0.223$ & $+1.55$ \\
        Relative height           & $+0.006$ & 0.454 & 0.456 & $+0.144$ & $-1.00$ \\
        Bounding-box area         & $-0.006$ & 0.205 & 0.202 & $-0.383$ & $+2.65$ \\
        Mask area                 & $+0.012$ & 0.103 & 0.103 & $+0.108$ & $-0.75$ \\
        Aspect ratio              & $-0.014$ & 1.250 & 1.208 & $+0.235$ & $-1.63$ \\
        Aspect distortion         & $+0.002$ & 0.136 & 0.138 & $-0.044$ & $+0.31$ \\
        Minimum-size constraint   & $-0.004$ & 0.102 & 0.101 & $+0.025$ & $-0.18$ \\
        Maximum-width constraint  & $-0.013$ & 0.689 & 0.654 & $-0.087$ & $+0.60$ \\
        Maximum-height constraint & $+0.010$ & 0.607 & 0.632 & $+0.199$ & $-1.38$ \\
        Both maximum constraints  & $-0.002$ & 0.310 & 0.301 & $-0.100$ & $+0.69$ \\
        \midrule
        \multicolumn{6}{@{}l}{\textit{Cutout geometry}} \\
        Outline complexity      & $-0.027$ & 6.875 & 6.743 & $+0.260$ & $-1.80$ \\
        Mask fill ratio         & $+0.022$ & 0.526 & 0.537 & $+0.362$ & $-2.51$ \\
        \bottomrule
    \end{tabular}
\end{table}

The measured properties vary little with semantic rank, with the largest absolute correlation reaching $0.027$. Their individual estimated contributions are also small relative to the observed 14.46-percentage-point decline. Bounding-box area gives the largest estimated decrease at $0.383$ percentage points, or 2.65\% of the observed change. These estimates describe individual associations rather than formal bounds and cannot be summed because the properties overlap.

\begin{figure}[!ht]
    \centering

    \begin{minipage}[c]{0.51\textwidth}
        \centering
        \footnotesize
        \setlength{\tabcolsep}{2pt}
        \begin{tabular}{@{}p{0.46\linewidth}rrr@{}}
            \toprule
            Specification &
            \shortstack{Median\\slope} &
            \shortstack{Retained\\(\%)} &
            \shortstack{CI\\below 0} \\
            \midrule
            Rank only
                & $-0.929$ & 100.0 & 35/35 \\
            + Placement and scale
                & $-0.979$ & 105.5 & 35/35 \\
            + Background effects
                & $-0.977$ & 104.4 & 35/35 \\
            + Object-instance effects
                & $-0.608$ & 66.5 & 35/35 \\
            Object-only alternative
                & $-0.782$ & 87.1 & 35/35 \\
            \bottomrule
        \end{tabular}
    \end{minipage}
    \hfill
    \begin{minipage}[c]{0.46\textwidth}
        \centering
        \includegraphics[width=\linewidth]{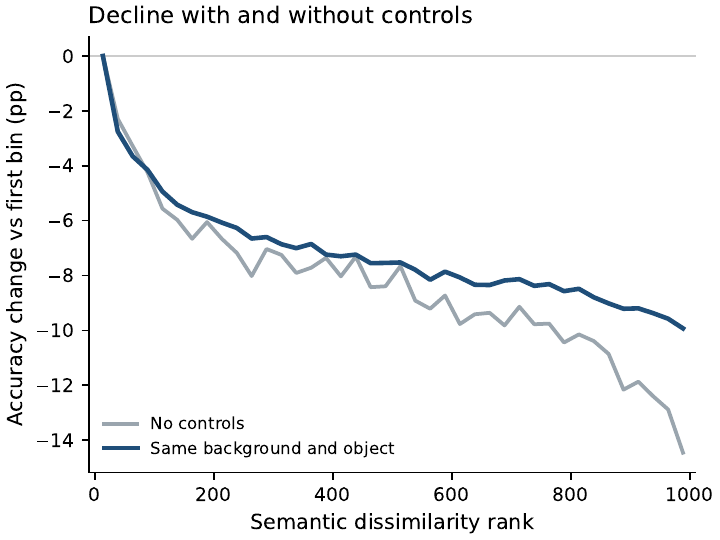}
    \end{minipage}

    \par\vspace{4pt}
    \noindent
    \begin{minipage}[t]{0.51\textwidth}
        \vspace{0pt}
        \captionof{table}{
        \textbf{Accuracy--distance slopes after adjustment.}
        Each $+$ row adds controls to those above it. The final row replaces object-instance effects with object-only accuracy, retaining the other controls. Slopes are medians across 35 models, in percentage points per 100 ranks. ``Retained'' is the median percentage of each model's rank-only slope remaining after adjustment. ``CI below 0'' counts models with entirely negative 95\% confidence intervals.
        }
        \label{tab:artifact_controls}
    \end{minipage}
    \hfill
    \begin{minipage}[t]{0.46\textwidth}
        \vspace{0pt}
        \caption{
        \textbf{Accuracy decline after adjustment.}
        Curves show accuracy changes relative to the first distance
        bin, averaged across all 35 models. The adjusted curve
        accounts for placement and scale measurements, background
        effects, and object-instance effects.
        }
        \label{fig:artifact_controls}
    \end{minipage}
\end{figure}

We next test whether the accuracy decline with semantic distance persists after accounting for object placement, size, and the difficulty of particular backgrounds and cutouts. The idea is to account for these properties when estimating the relationship between distance and accuracy. For each model, we fit a linear regression predicting top-1 correctness from semantic rank, then progressively add the controls shown in Table~\ref{tab:artifact_controls}. We first include 11 position and size predictors, comprising the two centre coordinates, eccentricity, border distance, sampled scale, mask area, aspect ratio, and the four size-constraint flags. We then account for some backgrounds being generally easier than others by estimating a separate baseline for each background. Finally, we add a separate baseline for each object instance to account for corresponding differences between cutouts. These background and object-instance fixed effects capture average differences in difficulty while retaining a common accuracy--distance slope. As an alternative to object-instance effects, we use each model's object-only accuracy for that cutout, averaged over black, white, and $1/f$ noise backgrounds, while retaining the preceding controls. Standard errors are clustered by background and object instance.

We also examine the shape of the relationship after adjustment. Rather than assuming that accuracy changes along a straight line, we divide semantic ranks into 40 bins of width 25 and estimate a separate adjusted accuracy change for each bin relative to the first. We retain the same position and size predictors and background and object-instance effects. The estimated changes are then averaged across all 35 models and compared with the unadjusted changes in observed accuracy.

All 35 models retain negative slopes with 95\% confidence intervals below zero under every specification (Table~\ref{tab:artifact_controls}). Placement and scale controls and background effects leave the slopes largely unchanged. Adding object-instance effects retains a median of 66.5\% of the original slope, while the object-only alternative retains 87.1\%. The adjusted curve also preserves the decreasing relationship (Figure~\ref{fig:artifact_controls}), with a 9.93-percentage-point decline compared with 14.46 points without controls and a Spearman correlation of $\rho=-0.993$ with bin index.

Finally, we test whether discontinuities at object boundaries explain the remaining decline. We reconstruct a separate sample of 30,000 composites covering 3,000 cutouts and measure luminance and colour differences across the object edge, together with local luminance gradients. Each measure is expressed relative to the corresponding region of the background alone. Adding these measures to the regression with placement and scale controls, background effects, and object-instance effects changes the median slope on this sample from $-0.585$ to $-0.583$ percentage points per 100 ranks. All 35 models retain confidence intervals below zero. Together, these controls show that the accuracy decline persists after accounting for measured generation properties and average differences between backgrounds and cutouts.

\FloatBarrier

\subsection{Object-Only Image Generation}
\label{sec:object-only-im_gen}

To isolate object-derived recognition signals, we construct object-only images by compositing segmented foreground objects onto neutral backgrounds. Foreground objects are extracted using the corresponding segmentation masks, which are binarized with a threshold of 127 to obtain a hard separation between foreground and background.

Each object is placed onto one of three background conditions: a uniform black background, a uniform white background, or a $1/f$ noise background. For the uniform conditions, all background pixels are set to a constant intensity value. For the $1/f$ noise condition, backgrounds are generated in the frequency domain by sampling random phase and scaling amplitudes by the inverse of spatial frequency in the amplitude spectrum, followed by an inverse Fourier transform. Noise is generated independently per channel using a deterministic seed derived from the image filename, ensuring reproducibility.

These images are used as controls in §\ref{sec:results_context_signal} and evaluated in Figure~\ref{fig:obj_bg_semantics}a, with full results across models reported in Appendix~\ref{sec:AppC1_Object_BG_Signal}. Example object-only images are shown in Figure~\ref{fig:dataset}a.

\subsection{Release, Licensing, and Ethical Considerations}
\label{sec:appendix_release}

ImageNet-OOC1k is derived from ImageNet validation images and ImageNet-S segmentation masks. Inpainted backgrounds are generated using \texttt{IOPaint} with the MAT inpainting model. Because the underlying images come from ImageNet, ImageNet-OOC1k inherits the access and redistribution constraints of ImageNet/ImageNet-S.

Code, metadata, semantic-distance matrices, and generation specifications are available at \url{https://github.com/ToastyDom/ImageNet-OOC1k}. Image-derived components are released subject to the access and redistribution terms of ImageNet/ImageNet-S. The accompanying documentation describes component provenance, access requirements, and inherited restrictions. Image-derived components, including backgrounds, object crops, masks, and composites, will be released in forms compatible with the access conditions and redistribution terms of the source datasets. We will not redistribute ImageNet-derived image content except in forms permitted by the source-data terms. The release documents the provenance of each component, access requirements, inherited restrictions, and the distinction between newly introduced assets and components derived from existing datasets.

ImageNet-OOC1k does not involve newly collected human-subject data, crowdsourcing, or participant interaction. The benchmark is intended for diagnostic evaluation of object--background context sensitivity in image recognition models, not for deployment decisions about people or safety-critical systems. Its main potential impact is to support more reliable evaluation of visual recognition models by exposing predictable context-dependent failures. At the same time, improved visual recognition can affect downstream applications whose societal consequences depend on the deployment context, including sensitive settings such as surveillance. The release will document intended use and known limitations, including the inherited scope and biases of ImageNet-style object categories.

\begin{figure}[!t]
    \centering
    \includegraphics[width=0.9\textwidth]{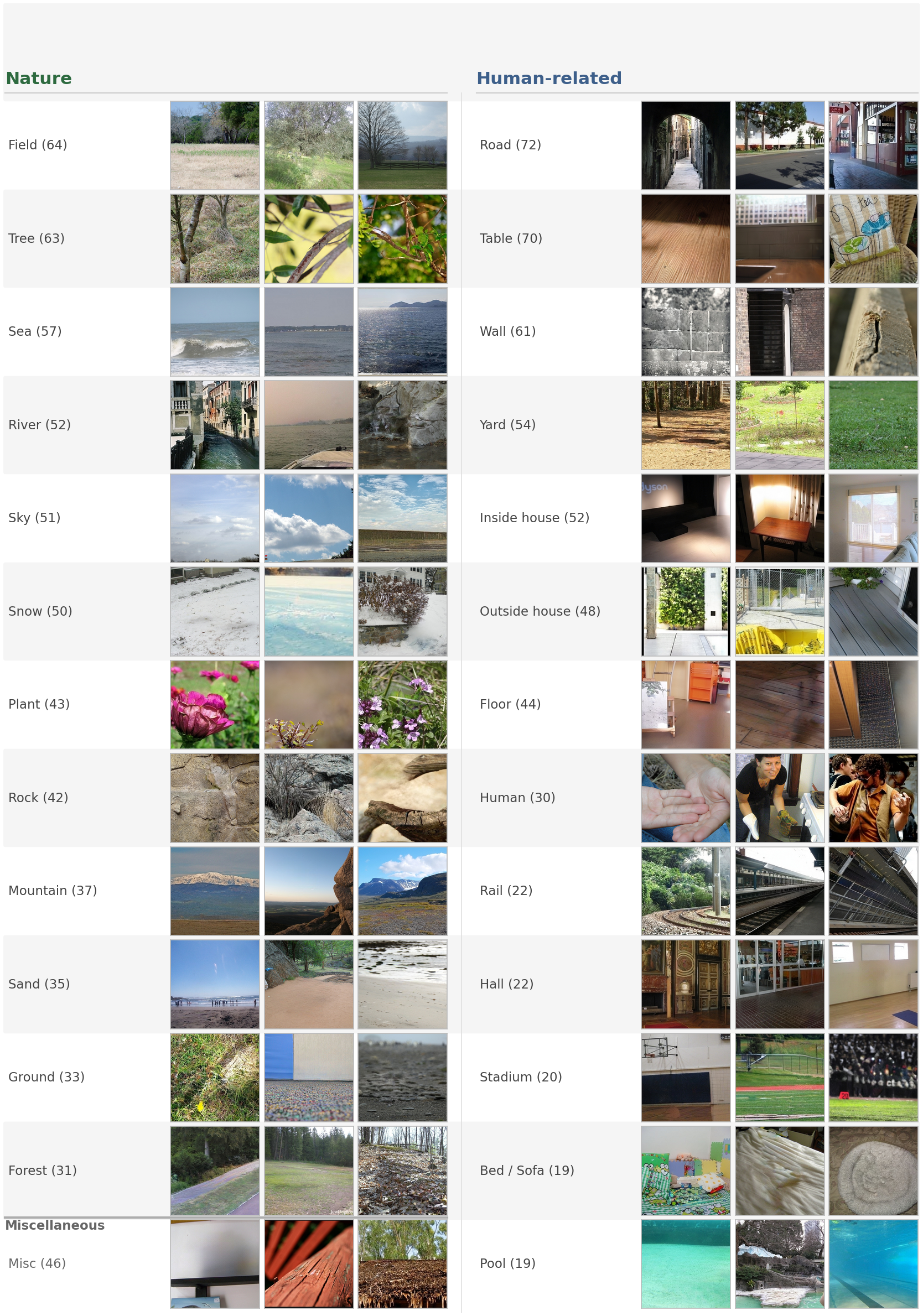}
    \caption{\textbf{Background scene subcategories in ImageNet-OOC1k.}
    Representative inpainted backgrounds organized by category and subcategory, with the number of scenes per subcategory shown in parentheses.
    The figure shows \emph{nature} environments (12 subcategories, 558 backgrounds), \emph{human-related} environments (13 subcategories, 533 backgrounds), and \emph{miscellaneous} scenes (46 backgrounds).
    All images show background-only scenes after removal of the original foreground object via inpainting. Subcategories are ordered by decreasing frequency within each category.}
    \label{fig:object_cats}
\end{figure}

\begin{table*}[t]
\centering
\caption{
\textbf{ImageNet-OOC1k dataset and composite generation summary.}
Counts are reported for the full dataset, the subset used in behavioral experiments (§\ref{Section:Semanticdistance}), and the subset used for mechanistic analysis (§\ref{sec:mechanistic}).
}
\label{tab:ooc_detailed}
\small
\begin{tabular*}{\textwidth}{@{\extracolsep{\fill}}lll}
\toprule
\textbf{Component} & \textbf{Detail} & \textbf{Count} \\
\midrule

\multicolumn{3}{l}{\textbf{Backgrounds}} \\
\midrule

\textbf{Nature} & field & 64 \\
 & tree & 63 \\
 & sea & 57 \\
 & river & 52 \\
 & sky & 51 \\
 & snow & 50 \\
 & plant & 43 \\
 & rock & 42 \\
 & mountain & 37 \\
 & sand & 35 \\
 & ground & 33 \\
 & forest & 31 \\
\textbf{Category total} & & \textbf{558} \\

\midrule

\textbf{Human-related} & road & 72 \\
 & table & 70 \\
 & wall & 61 \\
 & yard & 54 \\
 & inside\_house & 52 \\
 & outside\_house & 48 \\
 & floor & 44 \\
 & human & 30 \\
 & rail & 22 \\
 & hall & 22 \\
 & stadium & 20 \\
 & bed\_sofa & 19 \\
 & pool & 19 \\
\textbf{Category total} & & \textbf{533} \\

\midrule

\textbf{Misc} & -- & 46 \\
\textbf{Total backgrounds} & & \textbf{1,137} \\

\addlinespace
\midrule

\multicolumn{3}{l}{\textbf{Object Library}} \\
\midrule
Object image--mask pairs & Total & 12,321 \\
Object classes & ImageNet classes & 918 \\
 & Classes with at least four instances & 915 \\

\addlinespace
\midrule

\multicolumn{3}{l}{\textbf{Composites (Behavioral Experiments, §\ref{Section:Semanticdistance})}} \\
\midrule
Generated composites & Used in experiments & 953,862 \\
Backgrounds used & Stratified, 10 per background group & 260 \\
Possible pairings & Full object--background pool, upper bound & $\sim$14.0M \\

\addlinespace
\midrule

\multicolumn{3}{l}{\textbf{Subset (Mechanistic Analysis, §\ref{sec:mechanistic})}} \\
\midrule
Total composites & Transformer analysis & 75,000 \\
Sampling strategy & $20$ bins $\times$ $3$ categories $\times$ $1{,}250$/cell & -- \\

\bottomrule
\end{tabular*}
\end{table*}

\subsection{Compute Resources}
\label{app:compute}

All experiments were conducted on an internal university HPC cluster using a single NVIDIA A100-SXM4-80GB GPU per job, without multi-GPU parallelization. The evaluated vision models were not trained or fine-tuned. All experiments use publicly available pretrained weights loaded via \texttt{timm} (v1.0.26) and OpenAI CLIP. The software stack consisted of Python 3.11.15, PyTorch 2.10.0, CUDA 12.8, and cuDNN 9.10.02.

The compute reported here reflects the specific experimental sampling used in this paper, rather than an intrinsic requirement of the benchmark. In particular, we materialized and evaluated approximately 954K composite images to densely sample semantic distance across backgrounds and object classes. Batched inference across the 35 pretrained models used in the behavioral benchmark, together with 9 control conditions comprising approximately 55K additional images, required approximately 86 GPU-hours using a batch size of 128. The mechanistic analyses required additional extraction of patch-level logit-lens representations for the 75K-image stratified subset across five ViT variants: ViT-S/16, ViT-B/16, ViT-L/16, DeiT-B/16, and ViT-RelPos-B/16. ViT-RelPos-B/16 was included only in this mechanistic robustness analysis and is not part of the 35-model behavioral benchmark. This required approximately 11 GPU-hours.

Dataset generation was comparatively lightweight. Constructing the composite image set used in this paper required approximately 3.5 CPU-hours. Computing the consensus dissimilarity matrix from 10 pretrained classifiers took under 2 minutes on CPU. Background inpainting for the 1,137 source images was performed using \texttt{IOPaint}.

Storage requirements were modest and depend on the number of composites materialized. The ImageNet-OOC1k dataset instance used in this paper occupies approximately 98 GB, dominated by composite images stored in HDF5 format (86 GB). The remaining 12 GB includes original images, inpainted backgrounds, segmentation masks, and the object library. The consensus dissimilarity matrix and per-model matrices occupy 34 MB. Experimental outputs add approximately 10 GB, including prediction CSVs (5.6 GB) and patch-level activation files for five ViT variants (4.7 GB). Peak disk usage across the dataset and all experimental artifacts was approximately 109 GB.

In total, the reported experiments required approximately 97 GPU-hours and 4 CPU-hours. Preliminary experiments, including pipeline debugging, parameter tuning, and exploratory analyses, used the same inference-only setup and required an additional estimated 30--50 GPU-hours.

\FloatBarrier

\clearpage
\section{Semantic Structure of Classifier Representations}
\label{sec:appendix_semantic_structure}

Section~\ref{sec:consensus} defines the consensus visual-semantic axis from classifier-head representations. Here, we visualize the class-level geometry underlying this axis and compare it with individual model geometries. The goal is to show that the consensus matrix summarizes a visual-semantic structure already present across models, rather than combining unrelated class organizations.

For the consensus and selected individual models, we show the full class-level dissimilarity matrix and local neighborhoods for three representative classes (hen, holster, coral reef). Across supervised classifiers, semantically related classes form coherent local groups. Although fine-grained neighbor order varies, broad semantic neighborhoods recur across architectures, supporting the use of a consensus axis for the main experiments. For these classifiers, dissimilarities are computed from classifier-head representations as described in Section~\ref{sec:consensus}. For comparison, we encode ImageNet class-name prompts, \texttt{"a photo of a \{class\}"}, with the CLIP text encoder, L2-normalize the embeddings, and compute cosine dissimilarity.

\FloatBarrier

\subsection{Consensus Representation}

\begin{figure}[!h]
    \centering
    \includegraphics[width=\textwidth]{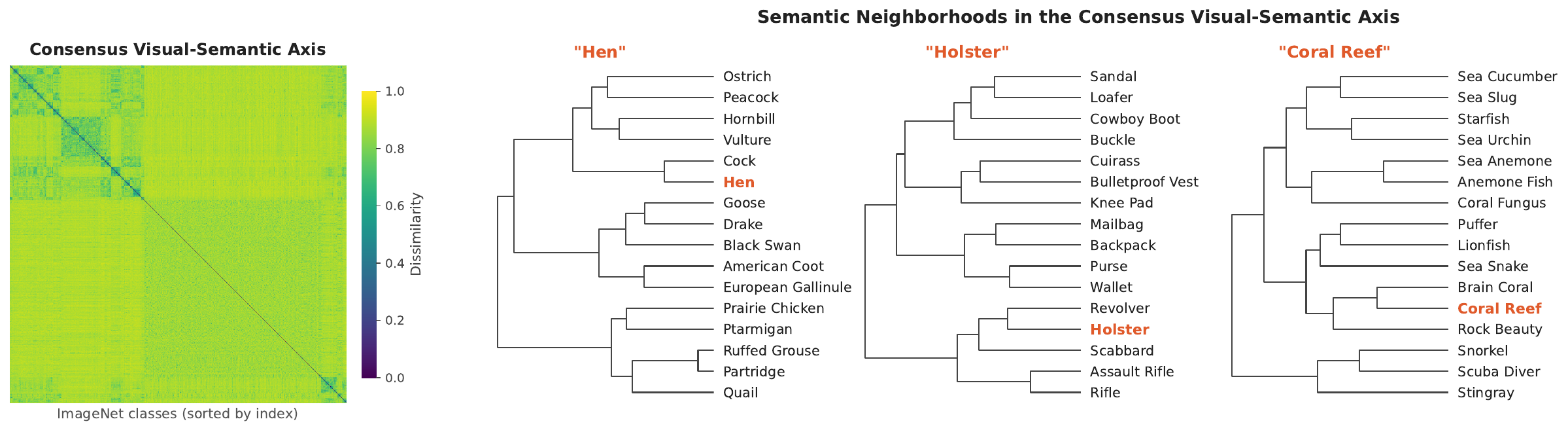}
    \caption{
    \textbf{Consensus visual-semantic structure.}
    Left: dissimilarity matrix over ImageNet classes derived from the consensus visual-semantic axis.
    Right: local neighborhoods for the same focal classes as in Figure~\ref{fig:dendrograms}, showing the 15 nearest neighbors clustered using Ward linkage.
    The consensus aggregates classifier-head representations across the ten pretrained models described in Section~\ref{sec:consensus} and defines the semantic axis used in the main experiments.
    }
    \label{fig:supp_consensus}
\end{figure}

\FloatBarrier

\subsection{Individual Model Representations}

\begin{figure}[!h]
    \centering
    \includegraphics[width=\textwidth]{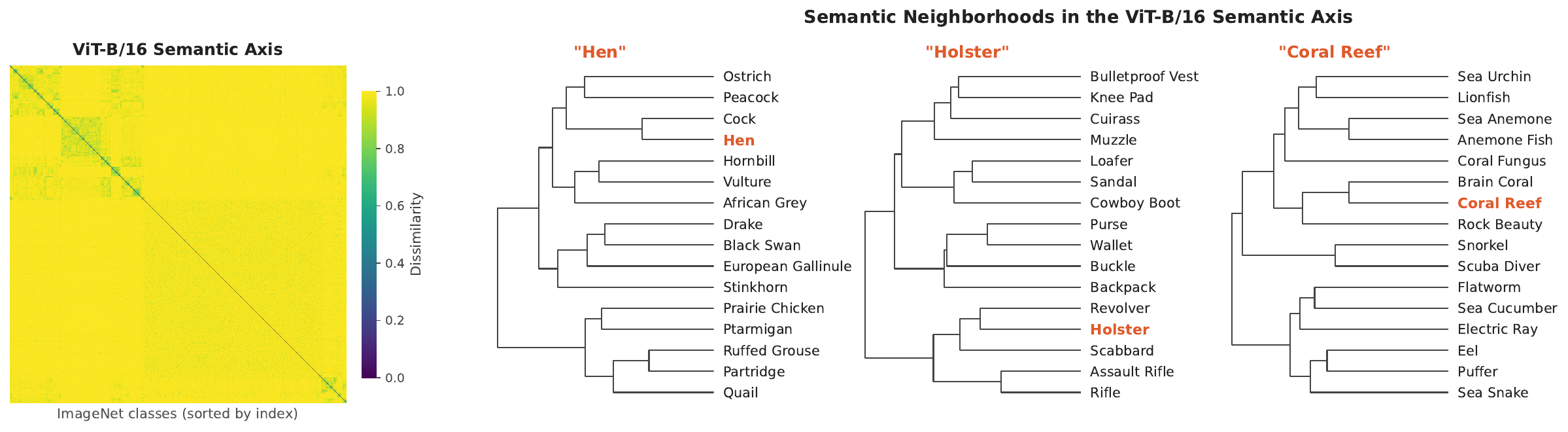}
    \caption{
    \textbf{Semantic structure for ViT-B/16.}
    Left: dissimilarity matrix over ImageNet classes derived from classifier-head representations.
    Right: local neighborhoods for the same focal classes as in Figure~\ref{fig:dendrograms}, showing the 15 nearest neighbors clustered using Ward linkage.
    Semantic groupings closely match the consensus structure, indicating that coherent class neighborhoods already emerge in an individual transformer model.
    }
    \label{fig:supp_vit}
\end{figure}

\FloatBarrier

\begin{figure}[!h]
    \centering
    \includegraphics[width=\textwidth]{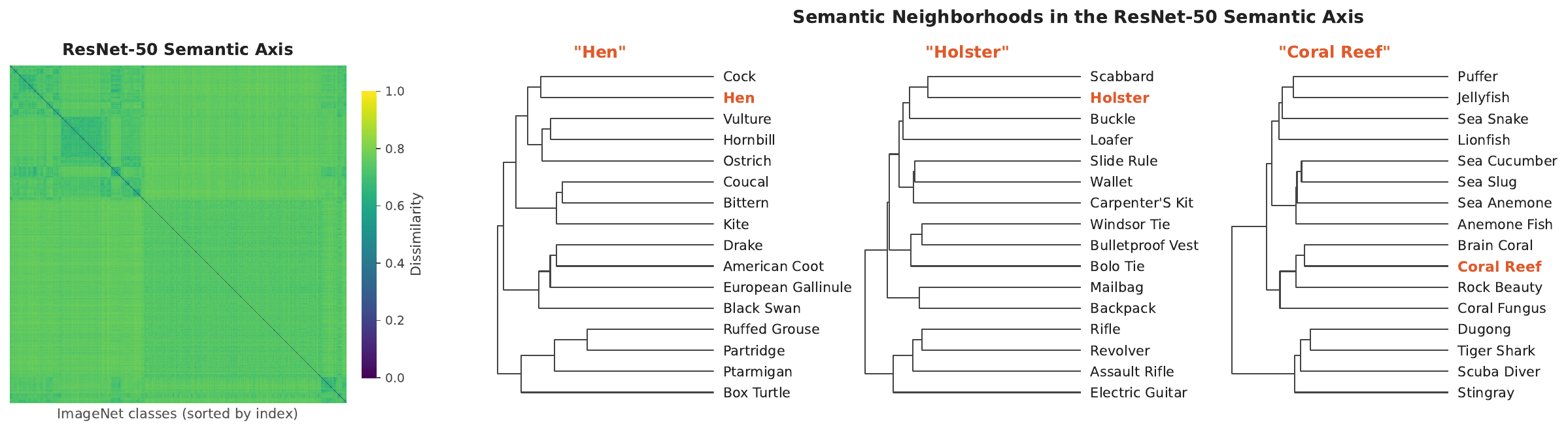}
    \caption{
    \textbf{Semantic structure for ResNet-50.}
    Left: dissimilarity matrix over ImageNet classes derived from classifier-head representations.
    Right: local neighborhoods for the same focal classes as in Figure~\ref{fig:dendrograms}, showing the 15 nearest neighbors clustered using Ward linkage.
    Similar semantic groupings emerge despite architectural differences, consistent with the consensus structure.
    }
    \label{fig:supp_resnet}
\end{figure}

\FloatBarrier

\begin{figure}[!h]
    \centering
    \includegraphics[width=\textwidth]{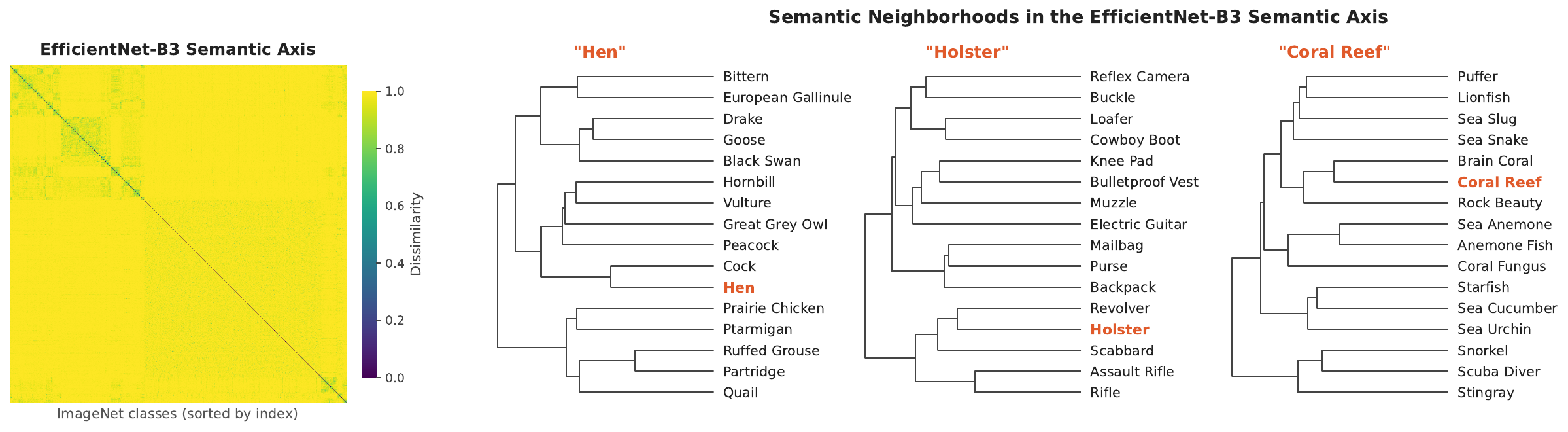}
    \caption{
    \textbf{Semantic structure for EfficientNet-B3.}
    Left: dissimilarity matrix over ImageNet classes derived from classifier-head representations.
    Right: local neighborhoods for the same focal classes as in Figure~\ref{fig:dendrograms}, showing the 15 nearest neighbors clustered using Ward linkage.
    High-level semantic organization is preserved, with variation primarily in fine-grained neighbor ordering.
    }
    \label{fig:supp_effnet}
\end{figure}

\FloatBarrier

\begin{figure}[!h]
    \centering
    \includegraphics[width=\textwidth]{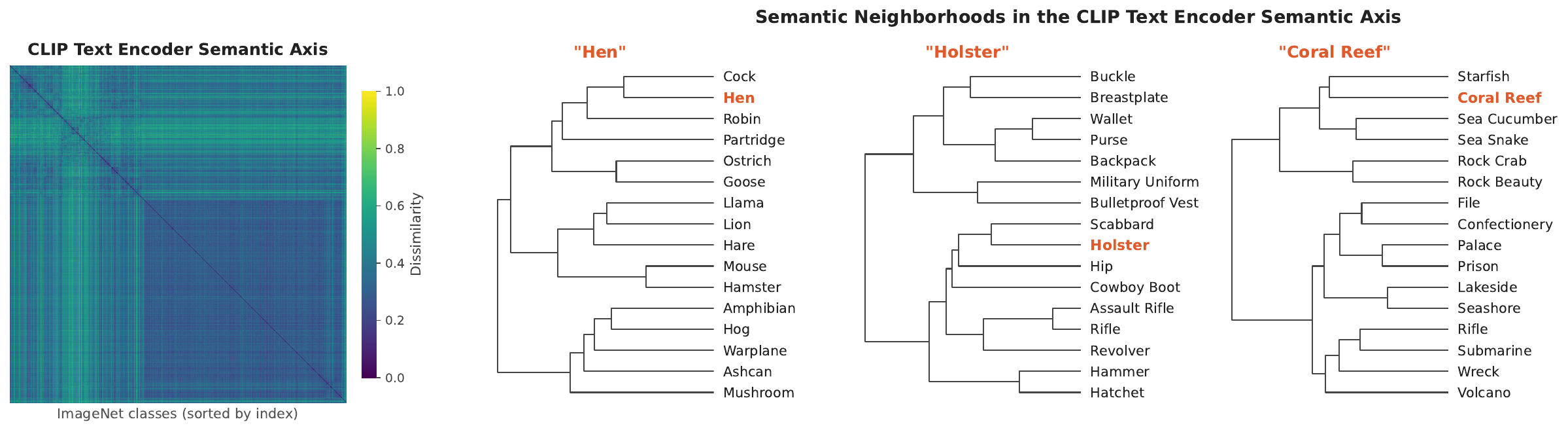}
    \caption{
    \textbf{Semantic structure for CLIP text embeddings.}
    Left: dissimilarity matrix over ImageNet classes derived from CLIP text encoder embeddings of class-name prompts.
    Right: local neighborhoods for the same focal classes as in Figure~\ref{fig:dendrograms}, showing the 15 nearest neighbors clustered using Ward linkage.
    The neighborhoods differ from those derived from classifier-head representations.
    }
    \label{fig:supp_clip}
\end{figure}

\FloatBarrier

\clearpage
\section{Robustness and Generalization of Behavioral Results}
\label{sec:appendix_bg_obj}

This section provides the robustness analyses supporting the behavioral results in Section~\ref{sec:5_Behavioral_Analysis}. The goal is to show that the main effects are not specific to the representative models shown in the main text, to a small subset of backgrounds, or to the construction of the consensus visual-semantic axis.

We first define the full model set used throughout the behavioral analyses (Table~\ref{tab:model_list}). We then extend the object-only and background-only controls from Section~\ref{sec:results_context_signal} to all models, showing that object- and background-derived signals remain separable across architectures. Next, we show that the monotonic accuracy decline with semantic distance holds across the full model set and examine how this global trend decomposes across individual backgrounds. Finally, we compare the consensus axis with alternative similarity spaces to verify that the observed effects are not a generic consequence of ranking classes by any semantic measure.

\begin{table*}[tbp]
\centering
\caption{
\textbf{Model set used in this study.}
Models are grouped into architecture families used throughout the behavioral analyses.
Models marked with $^\star$ correspond to the six representative models shown in the main-text behavioral figures.
}
\label{tab:model_list}
\small
\begin{tabular*}{\textwidth}{@{\extracolsep{\fill}}lll}
\toprule
\textbf{Model} & \textbf{Family} & \textbf{Reference} \\
\midrule

ViT-S/16 (vit\_small\_patch16\_224) & ViT & \cite{vit_paper} \\
ViT-B/16$^\star$ & ViT & \cite{vit_paper} \\
ViT-B/32 (vit\_base\_patch32\_224) & ViT & \cite{vit_paper} \\
ViT-B/16@384 (vit\_base\_patch16\_384) & ViT & \cite{vit_paper} \\
ViT-L/16$^\star$ & ViT & \cite{vit_paper} \\

\addlinespace
\midrule

DeiT-S/16 (deit\_small\_patch16\_224) & DeiT & \cite{Deit1} \\
DeiT-B/16$^\star$ & DeiT & \cite{Deit1} \\
DeiT-III-S (deit3\_small\_patch16\_224) & DeiT & \cite{Deit3} \\
DeiT-III-M (deit3\_medium\_patch16\_224) & DeiT & \cite{Deit3} \\
DeiT-III-B (deit3\_base\_patch16\_224) & DeiT & \cite{Deit3} \\
DeiT-III-L (deit3\_large\_patch16\_224) & DeiT & \cite{Deit3} \\

\addlinespace
\midrule

BEiT-B (beit\_base\_patch16\_224) & BEiT & \cite{bao2022beit} \\
BEiT-L (beit\_large\_patch16\_224) & BEiT & \cite{bao2022beit} \\
BEiT-v2-B (beitv2\_base\_patch16\_224) & BEiT & \cite{peng2022beitv2} \\
BEiT-v2-L (beitv2\_large\_patch16\_224) & BEiT & \cite{peng2022beitv2} \\
BEiT-3-B (beit3\_base\_patch16\_224) & BEiT & \cite{wang2023beit3} \\
BEiT-3-L (beit3\_large\_patch16\_224) & BEiT & \cite{wang2023beit3} \\

\addlinespace
\midrule

ResNet-34 & CNN & \cite{he2016resnet} \\
ResNet-50$^\star$ & CNN & \cite{he2016resnet} \\
ResNet-101 & CNN & \cite{he2016resnet} \\
ResNet-50 (SSL) & CNN & \cite{ssl_resnet} \\
Inception-v3 & CNN & \cite{szegedy2016inception} \\
Inception-v4 & CNN & \cite{szegedy2017inceptionv4} \\

\addlinespace
\midrule

EfficientNet-B2 & EfficientNet & \cite{tan2019efficientnet} \\
EfficientNet-B3$^\star$ & EfficientNet & \cite{tan2019efficientnet} \\
EfficientNet-B4 & EfficientNet & \cite{tan2019efficientnet} \\
EfficientNetV2-S & EfficientNet & \cite{tan2021efficientnetv2} \\
EfficientNetV2-M & EfficientNet & \cite{tan2021efficientnetv2} \\

\addlinespace
\midrule

ConvNeXt-S & ConvNeXt & \cite{liu2022convnext} \\
ConvNeXt-B & ConvNeXt & \cite{liu2022convnext} \\
ConvNeXt-L & ConvNeXt & \cite{liu2022convnext} \\
ConvNeXt-XL & ConvNeXt & \cite{liu2022convnext} \\
ConvNeXtV2-B & ConvNeXt & \cite{Woo_2023_CVPR} \\
ConvNeXtV2-L & ConvNeXt & \cite{Woo_2023_CVPR} \\

\addlinespace
\midrule

CLIP ViT-B/16$^\star$ & CLIP & \cite{clip_radford} \\

\bottomrule
\end{tabular*}
\end{table*}

\subsection{Object and Background Signals Across Models}
\label{sec:AppC1_Object_BG_Signal}

This section extends Figure~\ref{fig:obj_bg_semantics} to the full model set (Table~\ref{tab:model_list}). We evaluate both background-only and object-only controls under top-1 and top-5 accuracy to test whether the separation between background-derived and object-derived signals generalizes across architectures.

\begin{figure}[!h]
    \centering
    \includegraphics[width=0.95\textwidth]{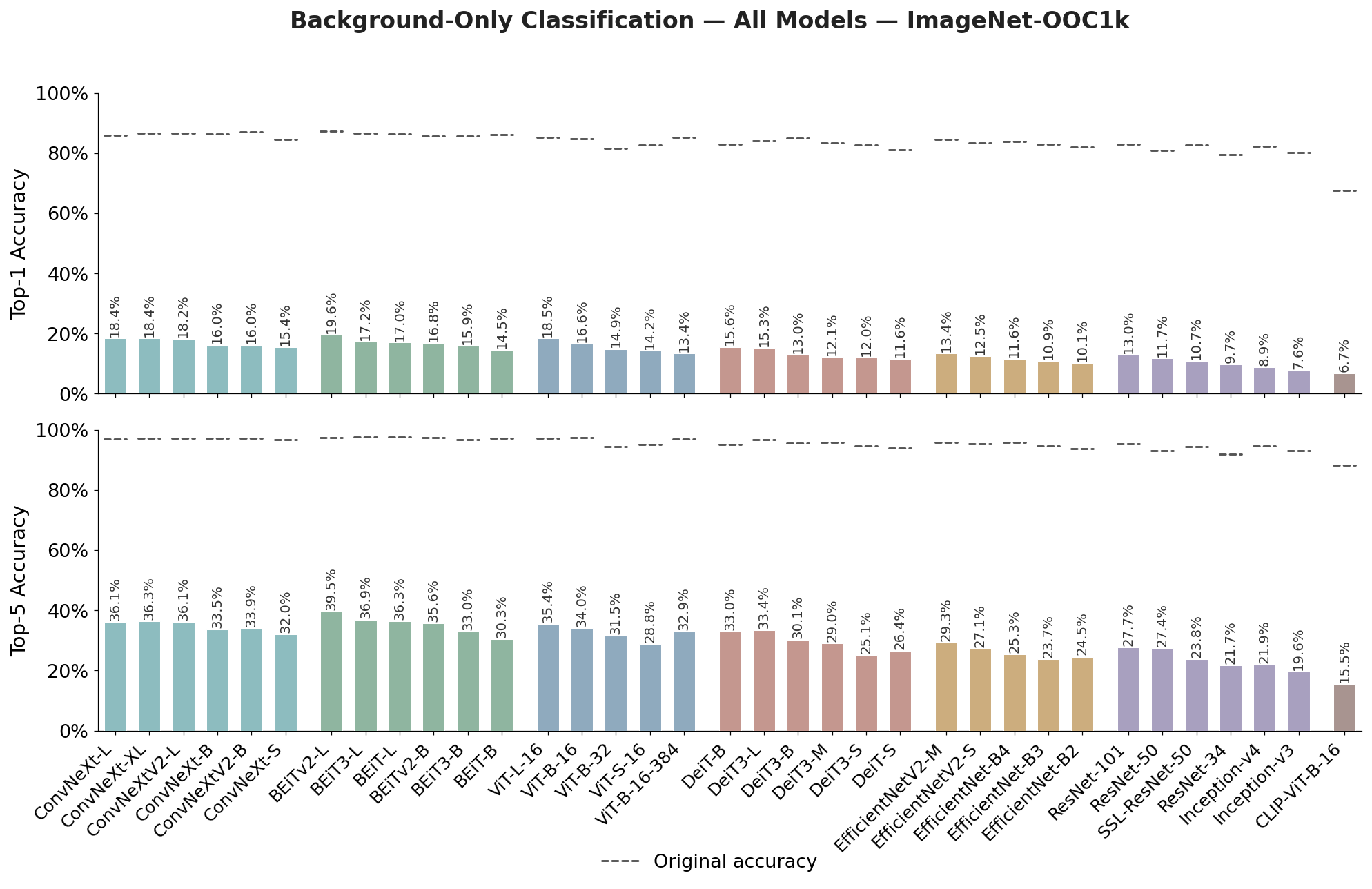}
    \caption{
    \textbf{Background-only accuracy across all models.}
    Rows show top-1 and top-5 accuracy for background-only images, with models grouped by architecture family and ordered by mean performance. All models perform above their respective chance baselines (0.1\% for top-1 and 0.5\% for top-5), showing that backgrounds alone carry class-specific signal. Top-5 accuracy increases substantially across models, indicating that the correct class often appears among the leading predictions even when the object is absent. The highest accuracies are observed for BEiT and ConvNeXt models, while CLIP and the evaluated CNNs show lower performance.
    }
    \label{fig:bg_all_comb}
\end{figure}

Background-only accuracy is consistently above chance (0.1\%) for all models under top-1 evaluation (Figure~\ref{fig:bg_all_comb}), confirming that backgrounds alone carry class-specific information. Although these backgrounds come from the ImageNet validation split, recurring object--context associations, such as aquatic animals with water, provide contextual cues that models can learn during training. Under top-5 evaluation, performance increases further, showing that the correct class frequently appears among the top predictions even when the object is absent. The highest background-only accuracies are observed for BEiT and ConvNeXt models, while CLIP and the evaluated CNNs show lower but still above-chance performance.

We next test whether the semantic-neighbor enrichment of incorrect top-1 predictions on background-only images could be explained by non-uniform class density in the consensus space. To do so, we computed a permutation null by shuffling object class labels across backgrounds while preserving the empirical distribution of model predictions. Across 1,000 permutations, the expected fraction of incorrect top-1 predictions falling within the top-25 semantic neighbors of the ground-truth class was 2.86\%, compared to 2.50\% under a uniform class baseline. The observed fraction was 25.48\%, corresponding to an 8.9-fold enrichment over the permutation null. None of the permutations reached the observed value ($p < 0.001$). This enrichment was consistent across models, ranging from 7.2-fold to 10.9-fold at $K=25$.

\begin{figure}[!h]
    \centering
    \includegraphics[width=\textwidth]{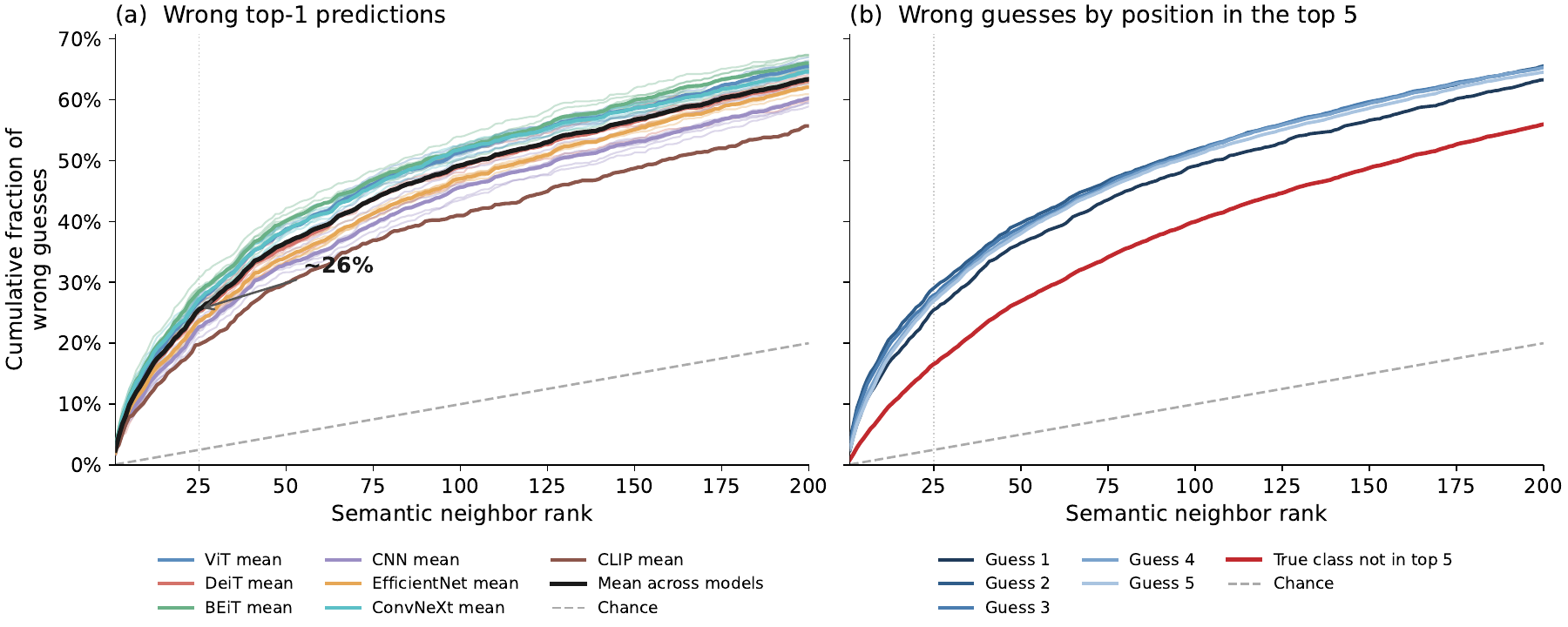}
    \caption{
    \textbf{Semantic structure of background-only predictions across all 35 models.}
    \textbf{(a)}~Cumulative visual-semantic neighbor-rank curves for incorrect top-1 predictions. Thin colored lines show individual models, thicker colored lines show architecture-family means, and the black line shows the mean across models.
    \textbf{(b)}~Corresponding curves for incorrect classes ranked first through fifth in the model predictions. The red curve pools all five predictions from images where the ground-truth class is absent from the top five.
    In both panels, the ground-truth class is excluded from the neighbor ranking, so rank 1 denotes its nearest other class. Dashed lines show the uniform baseline over the remaining 999 classes.
    }
    \label{fig:sem_struct_all}
\end{figure}

Figure~\ref{fig:sem_struct_all} shows that semantic-neighbor enrichment extends across architectures and prediction positions. For incorrect top-1 predictions (panel a), the mean fraction within the ground-truth class's 25 nearest visual-semantic neighbors is 25.6\%, with family means ranging from 19.8\% for CLIP to 28.5\% for BEiT. For incorrect classes at each of the top-five prediction positions (panel b), this fraction ranges from 25.5--29.0\%, compared with a uniform baseline of 2.5\%. Even when the ground-truth class is absent from the entire top five, 16.6\% of predicted classes fall within this neighborhood. Predictions outside these 25 neighbors can describe visible background content, such as \emph{seashore} for coastal scenes or \emph{valley} for mountains.

\begin{figure}[!h]
    \centering
    \includegraphics[width=0.95\textwidth]{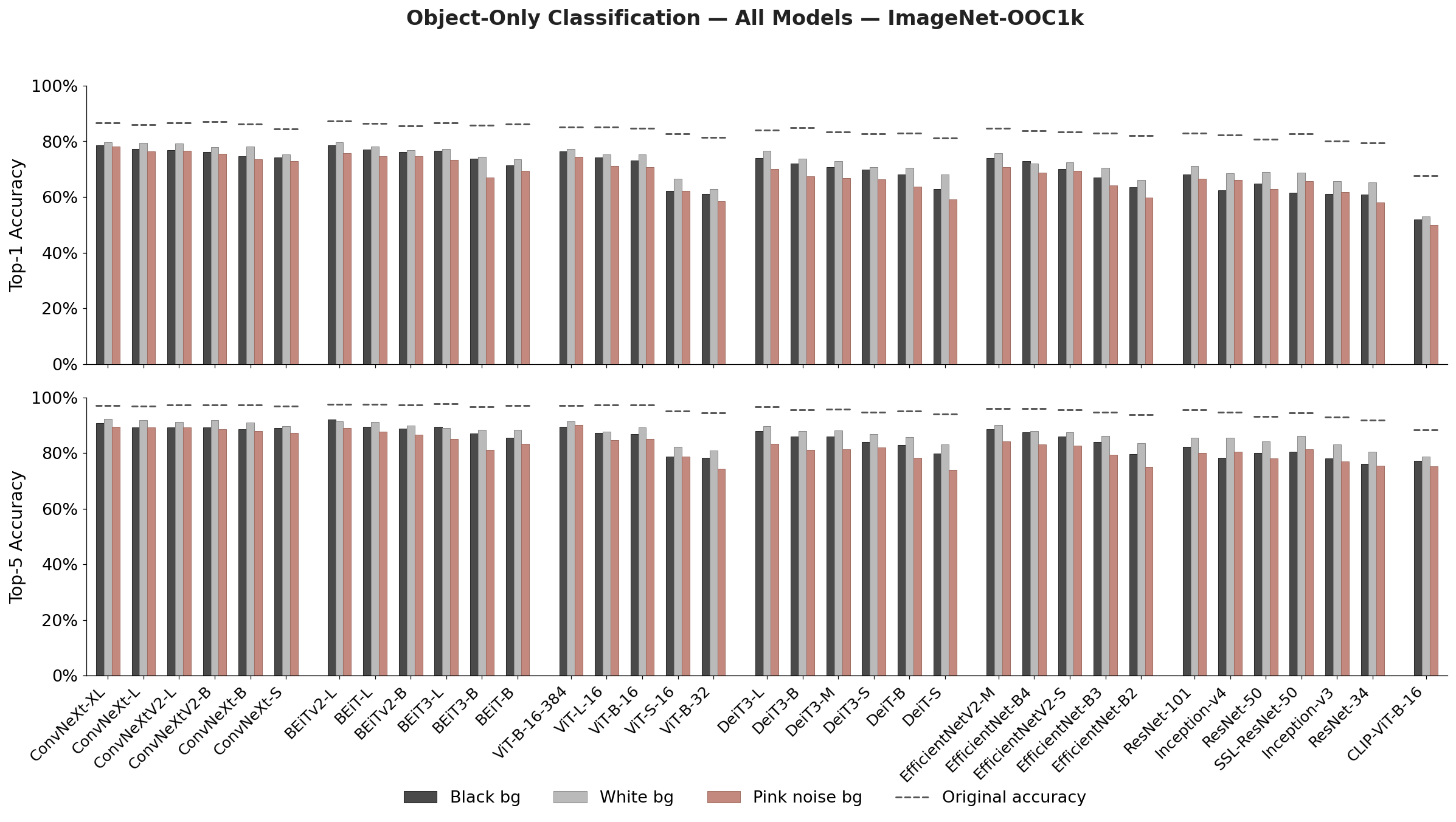}
    \caption{
    \textbf{Object-only accuracy across all models and neutral background conditions.}
    Rows show top-1 and top-5 accuracy for objects composited onto black, white, and $1/f$ noise backgrounds. Accuracy varies little across neutral background conditions in both evaluations, indicating that object-derived evidence is largely preserved under these manipulations. The same pattern holds across the full model set.
    }
    \label{fig:obj_all_comb}
\end{figure}

\newpage

\subsection{Universality Across Models}
\label{sec:appendix_semantic_distance}

\begin{figure}[!h]
\centering
\includegraphics[width=0.8\textwidth]{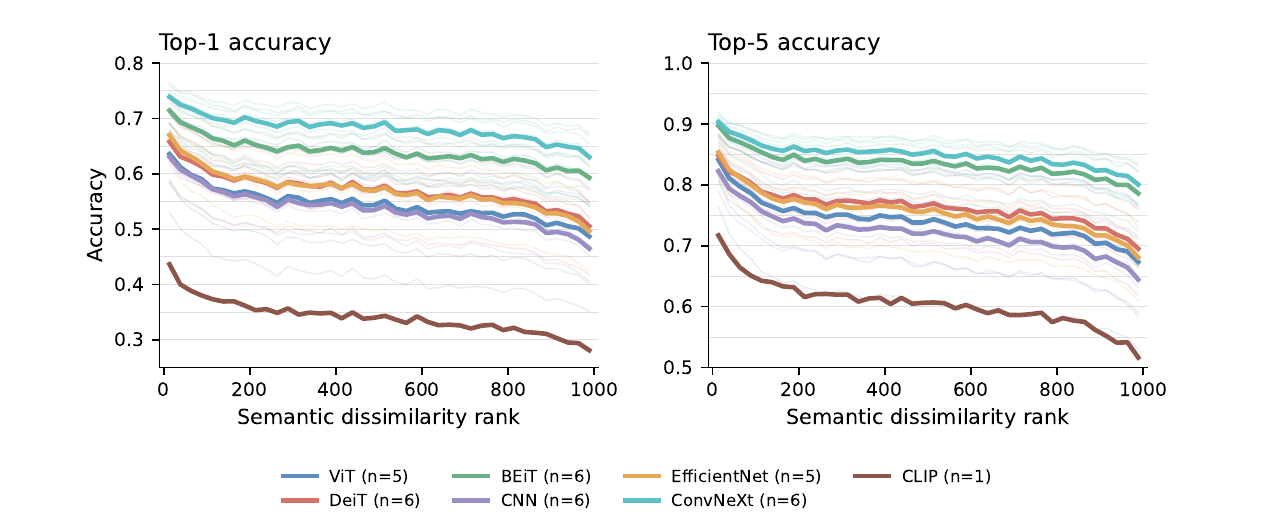}
\caption{
\textbf{Recognition accuracy as a function of semantic distance across all models.}
Thin lines show individual models and bold lines show architecture family means. All 35 models exhibit a consistent monotonic decline for both top-1 and top-5 accuracy as semantic object--background distance increases. Distances are grouped into bins of width 25 ranks for visualization.
}
\label{fig:appendix_all_models}
\end{figure}

To test whether the accuracy--distance pattern observed in Figure~\ref{fig:dataset}c generalizes beyond the representative models shown in the main text, we repeat the same analysis across the full 35-model set listed in Table~\ref{tab:model_list}. Figure~\ref{fig:appendix_all_models} shows that all models exhibit the same monotonic decline in recognition accuracy as semantic object--background distance increases. Although models differ substantially in baseline accuracy and in the magnitude of performance degradation, the functional form remains closely aligned across convolutional networks, vision transformers, and CLIP.

Across models, top-1 accuracy decreases by approximately 9--19 percentage points from the most similar to the most dissimilar composites. To verify that this trend is not an artifact of binned visualization, we fit an image-level logistic regression for each model, predicting correctness from semantic distance rank. All 35 models yielded strictly negative slopes, with all 95\% confidence intervals below zero (mean $\beta=-0.00038$ per rank; range $[-0.00048,-0.00028]$). This corresponds to an odds ratio of approximately $0.963$ per 100-rank increase, confirming that recognition probability decreases continuously with semantic distance at the image level.

\subsection{Heterogeneity Across Backgrounds}
\label{sec:appendix_het_across_bg}

To understand how the global accuracy--distance relationship arises, we analyze how it varies across individual backgrounds. For each of the 260 backgrounds, we compute a per-background interference score that summarizes the relationship between semantic distance and recognition accuracy. We collect all composites generated for a given background across the full range of semantic dissimilarity ranks and evaluate top-1 accuracy using five supervised models (ViT-B/16, ViT-L/16, DeiT-B/16, ResNet-50, and EfficientNet-B3). CLIP is excluded from this analysis because of its substantially lower baseline accuracy in this setting, although it shows the same qualitative trend. To reduce noise from sparse samples at individual ranks, we group ranks into bins of width 50, average accuracy within each bin across images and then across models, and compute the Spearman correlation $\rho$ between binned dissimilarity rank and accuracy. This yields one scalar interference score per background. Strongly negative values indicate decreasing accuracy with semantic distance, values near zero indicate no systematic effect, and positive values indicate facilitation, where accuracy increases with semantic distance.

The resulting scores form a broad distribution (mean $\rho = -0.32$, median $\rho = -0.49$, $\sigma = 0.52$). For visualization, we partition backgrounds into four descriptive bins using fixed thresholds on $\rho$, corresponding to strong negative ($\rho \leq -0.50$, $n = 129$), mid-negative ($-0.50 < \rho \leq 0.00$, $n = 55$), mid-positive ($0.00 < \rho \leq 0.50$, $n = 51$), and strong positive ($\rho > 0.50$, $n = 25$) relationships. These bins separate backgrounds with strong decreasing, weak decreasing, weak increasing, and strong increasing accuracy--distance trends.

\begin{figure}[!h]
\centering
\includegraphics[width=0.7\textwidth]{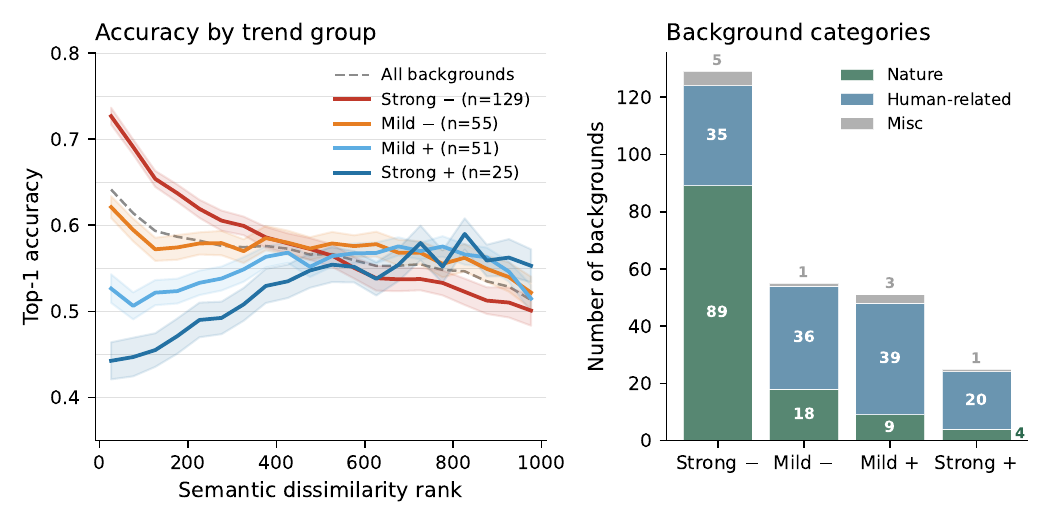}
\caption{
\textbf{Background-dependent modulation of semantic distance effects.}
\textbf{(a)}~Accuracy as a function of semantic distance, grouped by background-specific interference score $\rho$. Distances are grouped into bins of width 50 ranks for visualization. Shaded bands show $\pm 1$ standard error of the mean across models.
\textbf{(b)}~Distribution of background categories across interference groups. Most backgrounds show negative accuracy--distance trends, while a smaller subset exhibits weak or reversed effects.
}
\label{fig:appendix_interference}
\end{figure}

Most backgrounds exhibit negative correlations and contribute to the aggregate monotonic decline, while a smaller subset shows weak or positive correlations, corresponding to reduced or reversed sensitivity to semantic mismatch (Figure~\ref{fig:appendix_interference}). The monotonic decline therefore refers to accuracy averaged across backgrounds rather than a relationship that holds for every individual background. This variation is structured by scene type. Strong-interference backgrounds are predominantly nature scenes, whereas weak or positive effects occur more often in human-related environments.

The reasons for these background-specific differences remain unclear. An approximate grouping based on WordNet hypernyms shows that human-made and natural objects account for 49.3\% and 46.4\% of library images, respectively. Every background is paired with the same 918 object classes, so human-related backgrounds were not assigned a broader overall object pool than nature backgrounds. However, these observations do not establish which object--background pairings are plausible. Nature scenes may impose stronger expectations about which objects belong in them, whereas more heterogeneous human-related environments may support a broader range of plausible objects. This remains a hypothesis, and determining how object plausibility and scene diversity contribute to the observed trends requires further investigation.

\FloatBarrier
\newpage

\subsection{Validation of the Visual-Semantic Axis}
\label{sec:appendix_distance_validation}

This section supports the axis construction in Section~\ref{sec:consensus} and the accuracy--distance result in Section~\ref{Section:Semanticdistance}. To assess the robustness of our findings and address concerns about circularity, we examine whether the decline persists in non-contributing models and under axes derived from independently pretrained representations and alternative definitions of class similarity.

\paragraph{Model overlap.}
The consensus axis uses ten classifier heads, eight from evaluated models, which we call \emph{contributors}. If direct model overlap drives the observed pattern, we would expect particularly steep declines among contributors. We therefore use the same consensus axis for all 35 evaluated models and compare their mean first-to-last-bin accuracy declines. Contributors are compared with non-contributing siblings from the ResNet and EfficientNet families and models from families not represented in the consensus (Table~\ref{tab:axis_overlap}). Non-contributing siblings show the largest decline, while independent families also show a clear decline. The pattern therefore extends beyond contributors, and direct model overlap does not explain why the decline occurs across all 35 models.

\begin{table}[!ht]
\centering
\small
\begin{tabular}{lrr}
\hline
Evaluated-model group & $n$ & Mean decline (pp) \\
\hline
Axis contributors & 8 & 15.4 \\
Non-contributing siblings & 6 & 17.2 \\
Independent families & 21 & 13.3 \\
All models & 35 & 14.5 \\
\hline
\end{tabular}
\vspace{6pt}
\caption{
\textbf{Accuracy decline by participation in consensus construction.}
Decline is first-bin minus last-bin top-1 accuracy, using bins of width 25 ranks and averaged across models within each group. Siblings are non-contributing models from the ResNet and EfficientNet families. Independent families comprise DeiT, BEiT, ConvNeXt, Inception, and CLIP. EfficientNet-B5 and DenseNet-121 contribute to the consensus but are not among the evaluated models.
}
\label{tab:axis_overlap}
\end{table}

\paragraph{Own-axis comparisons.}
If direct reuse of a model's classifier head drives the effect, we would expect that model to decline more strongly along its own axis. We therefore keep each model's predictions fixed and re-rank the composites using each contributing classifier-head axis. We compare the decline obtained from the model's own head with its mean decline under other heads (Table~\ref{tab:axis_own}).

Among the seven contributors other than ResNet-101, only three decline more strongly along their own axis, and the mean difference is small at 0.3 percentage points. Their own classifier geometry therefore does not consistently produce a larger decline.

ResNet-101 appears to be an exception, declining by 24.3 percentage points along its own axis compared with 10.9 under other axes. However, applying the ResNet-101 axis to the other 34 evaluated models also produces a large mean decline of 20.5 percentage points. Its axis therefore produces steep declines for other models as well, so the large own-axis result alone does not demonstrate a selective match between ResNet-101 and its own classifier geometry.

\begin{table}[!ht]
\centering
\small
\begin{tabular}{lrrr}
\hline
Model & Own axis & Other axes & Difference \\
\hline
EfficientNet-B3 & 13.6 & 11.9 & $+1.7$ \\
ResNet-50 & 11.5 & 10.5 & $+1.0$ \\
ViT-B/16 & 9.8 & 8.8 & $+0.9$ \\
ViT-B/32 & 14.0 & 14.0 & $0.0$ \\
ViT-L/16 & 9.3 & 9.4 & $-0.1$ \\
ViT-S/16 & 13.6 & 14.2 & $-0.7$ \\
ViT-B/16@384 & 7.9 & 8.7 & $-0.9$ \\
Mean, excluding ResNet-101 & 11.4 & 11.1 & $+0.3$ \\
\hline
ResNet-101 & 24.3 & 10.9 & $+13.5$ \\
\hline
\end{tabular}
\vspace{6pt}
\caption{
\textbf{Own-axis versus other-axis accuracy decline.}
All values are first-bin minus last-bin top-1 accuracy in percentage points, using bins of width 25 ranks. Own-axis decline uses distances from the evaluated model's classifier head. Other-axis decline averages the declines obtained under other contributing heads. For the seven contributors other than ResNet-101, this mean excludes both the model's own head and the ResNet-101 head, averaging over eight axes. For ResNet-101, it averages over the other nine heads. Differences are computed before rounding.
}
\label{tab:axis_own}
\end{table}

\paragraph{Visual class representations.}
A remaining concern is that the original heads all come from supervised ImageNet classifiers. We therefore construct axes from frozen DINOv1 ViT-B/16, DINOv2 ViT-B/14, and MAE ViT-B/16 representations, none of which contributed to the consensus or appeared in the behavioral benchmark. DINOv1 and MAE were pretrained on ImageNet images without labels, while DINOv2 used the broader LVD-142M dataset, which includes ImageNet-derived images. For each backbone, we train a linear layer on frozen pre-logit features using ImageNet labels and cross-entropy loss. We train each probe for 20 epochs with AdamW, a learning rate of $10^{-3}$, and weight decay of $10^{-4}$. We then construct axes from the probe weight rows using the same procedure as for the original classifier heads. The three probe axes are also combined using the consensus procedure from Section~\ref{sec:consensus}.

\begin{figure}[!ht]
\centering
\includegraphics[width=\textwidth]{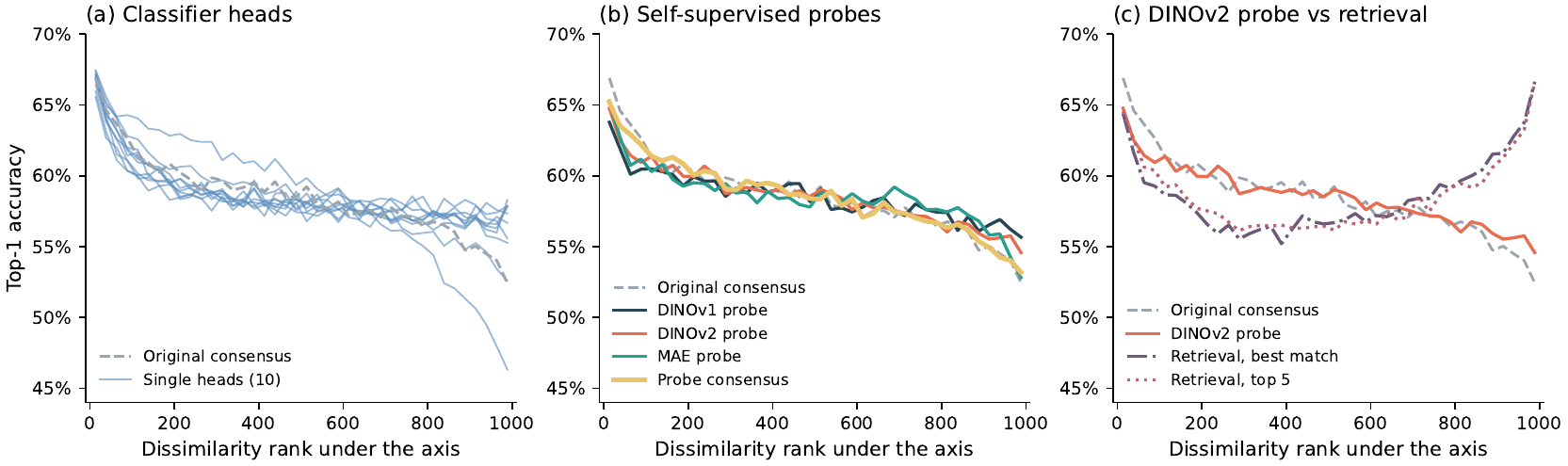}
\caption{
\textbf{Accuracy--distance relationships under visual classifier and retrieval axes.}
Each curve shows mean top-1 accuracy across the same 35 models and 953,862 composites, re-ranked under the given axis and grouped into bins of width 25. Dashed gray curves show the original consensus.
\textbf{(a)}~The ten contributing classifier-head axes, each shown as a thin blue curve.
\textbf{(b)}~Axes derived from linear probes on frozen DINOv1, DINOv2, and MAE representations, together with their probe consensus.
\textbf{(c)}~The DINOv2 probe axis compared with best-match and top-five retrieval axes derived from the same frozen representation. Probe axes reproduce the declining trend, while retrieval axes produce non-monotonic curves.
}
\label{fig:axis_validation_visual}
\end{figure}

All three independently constructed probe axes reproduce the declining trend (Figure~\ref{fig:axis_validation_visual}b). Their consensus yields a 12.0 percentage-point decline, compared with 14.5 for the original consensus. Across all 35 models, declines under the two consensuses correlate at $r=0.969$, despite only moderate agreement between the individual probe axes and the original consensus ($\rho=0.24$--$0.56$). Table~\ref{tab:axis_independent} reports the individual results and held-out probe accuracies. DINOv1 and DINOv2 provide the cleaner tests because the MAE probe has substantially lower held-out accuracy.

We also construct retrieval axes from the same frozen DINOv2 representation. Instead of learning to distinguish classes, retrieval asks how visually similar their image examples are. Using 50 images per class from the ImageNet training split, we compare each image with images from another class and record either its highest similarity or the average similarity of its five closest matches. We average these values over images to obtain class similarities, which are then converted to symmetric dissimilarities. Both retrieval axes produce U-shaped curves ending above their starting accuracy, unlike the declining probe curve (Figure~\ref{fig:axis_validation_visual}c).

\begin{table}[!ht]
\centering
\small
\begin{tabular}{lrrr}
\hline
Axis & Probe accuracy & Agreement $\rho$ & Decline (pp) \\
\hline
Original consensus & --- & 1.00 & 14.46 \\
DINOv1 probe & 74.7\% & 0.25 & 8.11 \\
DINOv2 probe & 81.2\% & 0.24 & 10.19 \\
MAE probe & 40.7\% & 0.56 & 12.46 \\
Probe consensus & --- & 0.50 & 12.00 \\
Retrieval, best match & --- & 0.43 & $-2.24$ \\
Retrieval, top five & --- & 0.43 & $-2.09$ \\
\hline
\end{tabular}
\vspace{6pt}
\caption{
\textbf{Accuracy--distance relationships under probe and retrieval axes.}
Probes are fitted using 90 images per class and evaluated on ten held-out images per class, all sampled from the ImageNet training split. Agreement is the Spearman correlation with the original consensus over unique off-diagonal class pairs. Decline is first-bin minus last-bin top-1 accuracy, averaged across all 35 models. Negative values indicate that accuracy ends above its starting value.
}
\label{tab:axis_independent}
\end{table}

\begin{figure}[!b]
\centering
\includegraphics[width=0.6667\textwidth]{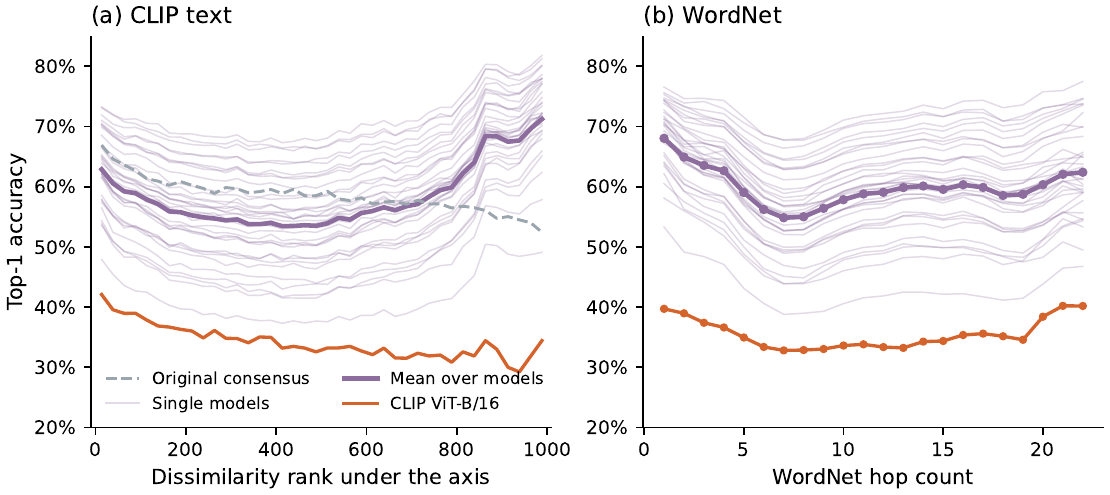}
\caption{
\textbf{Accuracy--distance relationships under linguistic and taxonomic similarities.}
Thin lines show individual evaluated models, the bold purple line shows the mean across all 35 models, and orange highlights CLIP ViT-B/16. Both panels use the same 953,862 composites.
\textbf{(a)}~Distances derived from CLIP text embeddings of ImageNet class-name prompts, grouped into bins of width 25 ranks. The dashed gray curve shows mean accuracy under the original consensus axis. Most models recover at larger CLIP-text distances, while CLIP itself retains a declining trend.
\textbf{(b)}~Accuracy by WordNet shortest-path hop count, showing only hop counts containing at least 1,000 composites. The relationship is U-shaped across models, including CLIP.
}
\label{fig:appendix_distance_validation}
\end{figure}

\paragraph{Linguistic and taxonomic similarities.}
Finally, we examine how recognition accuracy varies when class similarity reflects linguistic or taxonomic relationships (Figure~\ref{fig:appendix_distance_validation}). CLIP text embeddings produce a different curve from the visual classifier axes. Mean accuracy initially decreases but then rises above its starting value, with most individual models recovering toward the far end. CLIP itself retains an overall decline along its text axis, consistent with its joint image--text training. WordNet shortest-path distance yields a U-shaped relationship across individual models, including CLIP, rather than a consistent decrease.

Together, these controls demonstrate that the decline generalizes beyond the original consensus and contributing models, reproducing across independently pretrained discriminative representations. This convergence supports a shared visual-semantic structure relevant to recognition under object--background conflict. Classifier heads and probes capture directions that distinguish competing classes, directly addressing our question of which class the model selects when object and background evidence conflict. Retrieval, linguistic, and taxonomic measures describe other relationships between classes, so their different curves are compatible with this task-specific interpretation. These results support the consensus visual-semantic axis as a measure of shared discriminative visual class structure over ImageNet categories. Its alignment with human similarity judgments remains to be tested.

\section{Spatial Diagnostics and Recovery of Aggregation Failures}
\label{sec:appendix_recovery}

This section provides the definitions and additional analyses supporting the aggregation-failure results in Section~\ref{sec:mechanistic}. We first define the core metrics used throughout the analysis, including semantic pull, realm balance, and classification proximity. We then define the outcome taxonomy used to separate clean wins, contested wins, found-but-lost cases, and no-detection cases. Finally, we characterize the spatial structure of found-but-lost errors and describe the re-ranking strategy, operating regimes, and recovery results shown in Figure~\ref{fig:fbl_recovery}.

\subsection{Core Metrics}
\label{Coremetrics}

We use three core metrics to track how object- and background-related evidence evolves across layers. Semantic pull measures whether a patch prediction is closer to the object class or the background class. Realm balance summarizes the fraction of decided patches favoring the object. Classification proximity measures how close the CLS prediction is to the true object class in the consensus space.

For each patch token $p$ at layer $\ell$, we obtain a logit-lens prediction $\hat{c}_{p}^{(\ell)}$. We define \textbf{semantic pull} as
\[
s_{p}^{(\ell)}
=
D^{\mathrm{cons}}\!\left(\hat{c}_{p}^{(\ell)}, c_{\mathrm{bg}}\right)
-
D^{\mathrm{cons}}\!\left(\hat{c}_{p}^{(\ell)}, c_{\mathrm{obj}}\right),
\]
where $D^{\mathrm{cons}}$ is the consensus dissimilarity matrix, $c_{\mathrm{obj}}$ is the ground-truth object class, and $c_{\mathrm{bg}}$ is the class originally associated with the background. Positive values indicate predictions closer to the object class than to the background class, while negative values indicate predictions closer to the background class.

A patch is considered decided if $|s_{p}^{(\ell)}| > \tau$. We set $\tau = 0.0945$, corresponding to the mean standard deviation of semantic pull across early layers (blocks 0--7), where patch predictions are uninformative. Object-realm patches satisfy $s_{p}^{(\ell)} > \tau$, and background-realm patches satisfy $s_{p}^{(\ell)} < -\tau$.

For image $i$, we define \textbf{realm balance} as the fraction of decided patches favoring the object,
\[
\mathrm{RB}_{i}^{(\ell)}
=
\frac{
|\{p : s_{p}^{(\ell)} > \tau\}|
}{
|\{p : s_{p}^{(\ell)} > \tau\}|
+
|\{p : s_{p}^{(\ell)} < -\tau\}|
}.
\]
Values above 0.5 indicate that more decided patches favor the object than the background, while values below 0.5 indicate background dominance.

For the CLS token, we define \textbf{classification proximity} to the object class as
\[
\mathrm{CP}^{(\ell)}
=
1 -
D^{\mathrm{cons}}\!\left(\hat{c}_{\mathrm{CLS}}^{(\ell)}, c_{\mathrm{obj}}\right),
\]
where $\hat{c}_{\mathrm{CLS}}^{(\ell)}$ is the logit-lens prediction of the CLS token at layer $\ell$. Higher values indicate that the CLS prediction is closer to the true object class in the consensus space.

\subsection{Outcome Taxonomy}
\label{sec:appendix_taxonomy}

We define a four-way outcome taxonomy to characterize model behavior at the image level. All definitions are evaluated at post-norm. For each image $i$, let $e_i$ denote the number of patches (out of 196) whose argmax prediction exactly matches the object class, and let $\mathrm{RB}_i$ denote realm balance at post-norm.

Each image is assigned to one of four mutually exclusive categories:
\begin{itemize}
    \item \textbf{Clean win:} correct prediction with $e_i \geq 20$ and $\mathrm{RB}_i > 0.75$
    \item \textbf{Contested win:} correct prediction but not a clean win
    \item \textbf{Found-but-lost:} incorrect prediction with $e_i \geq 20$
    \item \textbf{No detection:} incorrect prediction with $e_i < 20$
\end{itemize}

The detection threshold $k = 20$ captures 98.9\% of correct images; the remaining 1.1\% of correct cases with $e_i < 20$ are included in the contested win category. Results are qualitatively stable for $k \in [20, 30]$.

\begin{figure}[!t]
\centering
\includegraphics[width=\textwidth]{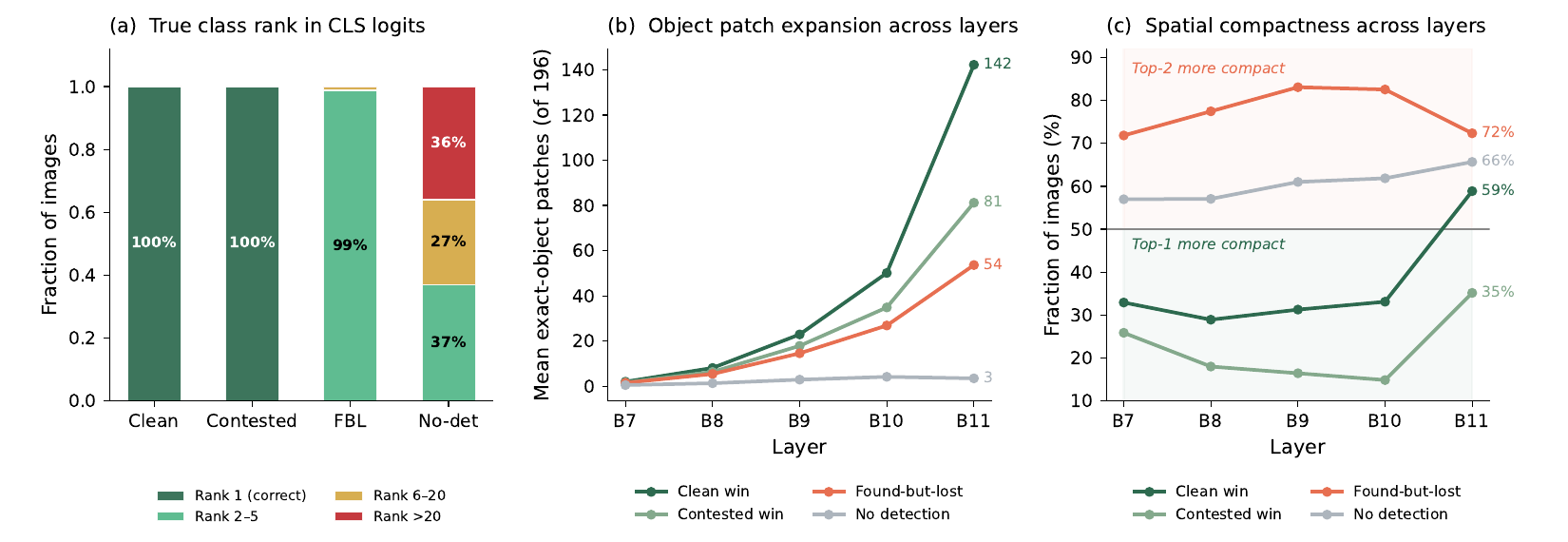}
\caption{
\textbf{Diagnostic signatures of aggregation failure.}
All panels use the 75,000-composite stratified subset analyzed for ViT-B/16, split by outcome category.
\textbf{(a)}~Distribution of the true object-class rank in post-norm CLS logits. FBL cases are near-miss errors, with the true class typically ranked close to the final prediction.
\textbf{(b)}~Mean number of patches whose logit-lens argmax matches the object class across transformer blocks. Clean wins show broad expansion of object-aligned patches, FBL cases show partial expansion, and no-detection cases remain near baseline.
\textbf{(c)}~Fraction of images where the runner-up CLS class (top-2) is more spatially compact than the predicted CLS class (top-1). Values above 0.5 indicate that top-2 is more compact more often, while values below 0.5 indicate that top-1 is more compact more often. Compactness is measured as the mean distance of supporting patches from their centroid. FBL cases show a strong top-2 compactness advantage at intermediate layers, indicating a localized object signal that has not been fully aggregated.
}
\label{fig:appendix_FBL_fix}
\end{figure}

\subsection{Diagnostic Signatures of Aggregation Failure}
\label{app:Diagnosticsignatures}

Figure~\ref{fig:appendix_FBL_fix} extends the analysis in Section~\ref{sec:agg_failure} by characterizing patterns in found-but-lost (FBL) cases that are consistent with aggregation limitations. All analyses are performed on ViT-B/16 using the logit lens on 75,000 composite images, with outcomes defined at post-norm using the taxonomy in Section~\ref{sec:appendix_taxonomy}.

Panel (a) shows that FBL cases are near-miss errors. For each image, we examine the rank of the true object class in the CLS logits at post-norm. While clean and contested wins are rank 1 by definition, FBL cases typically keep the correct class highly ranked despite the incorrect final prediction. The true class lies within the top-5 predictions in 98.7\% of FBL cases, with median rank 2, compared to 37\% and median rank 9 for no-detection cases. This confirms that FBL cases often contain the correct answer in the CLS representation, but fail to select it.

Panel (b) shows that this near-miss structure is accompanied by partial spatial expansion. Clean-win cases expand from approximately 2 to 142 exact object-class patches across layers, indicating that object evidence spreads broadly across the grid before the final prediction. FBL cases also detect the object, but expand only to around 54 patches. No-detection cases remain near baseline, increasing only from approximately 0.5 to 3.5 patches. Thus, the difference is not simply whether the object is found, but whether object evidence expands enough to dominate the spatial representation.

Panel (c) asks whether incomplete expansion leaves a recoverable spatial signature. The intuition is that object evidence in FBL cases may remain spatially localized, forming a compact cluster of supporting patches, while the competing predicted class is more spatially dispersed. We measure spatial spread as the mean Euclidean distance of supporting patches from their centroid on the $14 \times 14$ grid, as formalized in Section~\ref{app:reranking}. Lower spread indicates a more compact class-specific patch cluster. The plot reports the fraction of images in which the runner-up class (top-2) is more compact than the predicted class (top-1). Values above 0.5 therefore indicate that the runner-up is more compact more often, while values below 0.5 indicate that the predicted class is more compact more often. In FBL cases, the runner-up is often the true object class, and it is more compact than the predicted class at intermediate layers, especially blocks 9--10. At block 10, this occurs in approximately 82\% of FBL cases, compared to about 33\% of clean wins. Clean-win cases show the opposite pattern, with the predicted class forming the more compact cluster. Thus, FBL cases contain a compact and localized object signal that has not been selected by the CLS token. The gap collapses at post-norm to roughly 13 percentage points, 72\% versus 59\%, as representations become fully aggregated. This identifies block 10 as a stage where aggregation is still unresolved and object evidence remains recoverably structured.

\section{Spatial Re-Ranking as a Recoverability Probe}
\label{app:reranking}

To test whether FBL errors contain recoverable structure, we evaluate simple re-ranking rules that swap the CLS top-1 prediction with the top-2 prediction under different gating conditions. This is intended as a diagnostic probe, not as a deployment-time correction method. Although the rules are designed around FBL cases, a swap can also incidentally correct other errors or break originally correct predictions. Table~\ref{tab:recovery_gates} reports these tradeoffs.

\paragraph{Spatial spread.}
Let $\hat{c}_1$ and $\hat{c}_2$ denote the top-1 and top-2 classes from the post-norm CLS logits. For a candidate class $c$ at transformer block $\ell$, we define its patch support as the set of patches whose logit-lens prediction equals $c$,
\[
\mathcal{P}_c^{(\ell)} = \{p : \hat{c}_p^{(\ell)} = c\}.
\]
The spatial spread of $c$ is the mean Euclidean distance of its supporting patches from their centroid on the $H \times W$ patch grid,
\[
\mathrm{spread}(c,\ell) =
\frac{1}{|\mathcal{P}_c^{(\ell)}|}
\sum_{p \in \mathcal{P}_c^{(\ell)}}
\left\| \mathbf{x}_p - \bar{\mathbf{x}}_c^{(\ell)} \right\|_2,
\qquad
\bar{\mathbf{x}}_c^{(\ell)} =
\frac{1}{|\mathcal{P}_c^{(\ell)}|}
\sum_{p \in \mathcal{P}_c^{(\ell)}} \mathbf{x}_p ,
\]
where $\mathbf{x}_p=(r_p, q_p)$ denotes the patch-grid coordinates. If $|\mathcal{P}_c^{(\ell)}|=0$, spread is set to the maximum grid diagonal, $\sqrt{(H-1)^2 + (W-1)^2}$. If $|\mathcal{P}_c^{(\ell)}|=1$, spread is set to zero.

\paragraph{Gating signals.}
Each rule combines up to three signals motivated by the FBL signature described in Section~\ref{sec:agg_failure}. The spatial compactness advantage $a$ is measured at the penultimate transformer block $\ell_{\mathrm{pen}}$, block 10 for ViT-B/16,
\[
a =
\mathrm{spread}(\hat{c}_1,\ell_{\mathrm{pen}})
-
\mathrm{spread}(\hat{c}_2,\ell_{\mathrm{pen}}).
\]
Positive values indicate that the runner-up class forms a tighter patch cluster than the top-1 class. The logit gap is
\[
g = z_{\hat{c}_1} - z_{\hat{c}_2},
\]
where $z_{\hat{c}_1}$ and $z_{\hat{c}_2}$ are the post-norm CLS logits. Smaller values indicate a more uncertain final decision. The semantic distance is
\[
d = D^{\mathrm{cons}}_{\hat{c}_1,\hat{c}_2},
\]
computed from the consensus dissimilarity matrix.

\paragraph{Re-ranking rules.}
A swap replaces the top-1 prediction with the top-2 prediction when the selected gating conditions are satisfied. Unprotected rules apply the swap when
\[
a > a_{\min}
\qquad \text{and} \qquad
g < g_{\max}.
\]
Protected rules add a semantic-distance constraint,
\[
d > d_{\min},
\]
to suppress near-neighbor swaps and reduce breakage.

\paragraph{Results.}
The threshold settings in Table~\ref{tab:recovery_gates} illustrate the recovery--breakage tradeoff. Stricter unprotected settings yield high precision but recover fewer FBL cases, while relaxed settings recover more FBL cases at the cost of more broken correct predictions. Adding the semantic-distance constraint improves this tradeoff. The selected rule, $a > 1.0$, $g < 1.0$, and $d > 0.5$, recovers 4,277 FBL cases, incidentally corrects 428 non-FBL errors, and breaks 456 correct predictions. This yields a 9.4:1 FBL recovery-to-breakage ratio and a net gain of +5.7 percentage points over the 75,000-image ViT-B/16 diagnostic subset. Since thresholds are selected on the same subset used for the diagnostic analysis, this gain should be interpreted as an in-sample estimate of recoverability. Increasing $d_{\min}$ further improves the recovery-to-breakage ratio while preserving a similar net gain.

Together, these results show that FBL errors leave a structured spatial signature that can be detected from intermediate representations. The recovery analysis supports the interpretation that, in this subset of failures, object information is present but not selected by the final aggregation step.

\FloatBarrier

\begin{table}[!h]
\centering
\caption{\textbf{Re-ranking strategies and performance tradeoffs.}
Each rule specifies thresholds on spatial compactness advantage ($a$), logit gap ($g$), and optionally semantic distance ($d$). The swap from the top-1 to the top-2 prediction is applied only when all conditions are satisfied. ``FBL recovered'' denotes found-but-lost errors corrected by the swap, ``Other corrected'' denotes non-FBL errors corrected incidentally, and ``Broken'' denotes originally correct predictions made incorrect. ``Ratio'' is computed as ``FBL recovered'' divided by ``Broken''. Net $\Delta$ is computed over all 75,000 images. See Figure~\ref{fig:fbl_recovery} for the corresponding visualization.}
\label{tab:recovery_gates}
\small
\begin{tabular}{lccccc}
\toprule
\textbf{Rule} & \textbf{FBL recovered} & \textbf{Other corrected} & \textbf{Broken} & \textbf{Ratio} & \textbf{Net $\Delta$} \\
\midrule
$a > 2.0$, $g < 0.5$ & 1,786 & 171 & 185 & 9.7:1 & +2.4pp \\
$a > 1.0$, $g < 1.0$ & 4,523 & 518 & 953 & 4.7:1 & +5.5pp \\
$a > 1.0$, $g < 1.5$ & 5,601 & 833 & 1,574 & 3.6:1 & +6.5pp \\
\textbf{$\star$ $a > 1.0$, $g < 1.0$, $d > 0.5$} & \textbf{4,277} & \textbf{428} & \textbf{456} & \textbf{9.4:1} & \textbf{+5.7pp} \\
$a > 1.0$, $g < 1.0$, $d > 0.6$ & 4,184 & 397 & 341 & 12.3:1 & +5.7pp \\
$a > 1.0$, $g < 1.0$, $d > 0.7$ & 4,089 & 363 & 265 & 15.4:1 & +5.6pp \\
\bottomrule
\end{tabular}
\end{table}

\FloatBarrier


\section{Cross-Architecture Robustness of Aggregation-Failure Diagnostics}
\label{sec:appendix_cross_model}

This section tests whether the aggregation-failure diagnostics identified in ViT-B/16 are specific to that architecture or to CLS-token pooling. We repeat the core analyses across four additional transformer variants: ViT-S/16, ViT-L/16, DeiT-B/16, and ViT-RelPos-B/16~\cite{rw2019timm, liu2021swin}. ViT-RelPos-B/16 uses average pooling rather than CLS-token pooling, allowing us to test whether similar signatures arise when the final prediction is computed from pooled patch representations rather than a dedicated aggregation token. For this model, we compute the top-1 and top-2 classes from the mean of the final-layer patch logits. FBL cases therefore identify errors in which object-aligned patches are present but the averaged logits favor another class.

\subsection{Outcome Composition Across Architectures}
\label{app:tax_gen}

\begin{figure}[!h]
    \centering
    \includegraphics[width=0.6\textwidth]{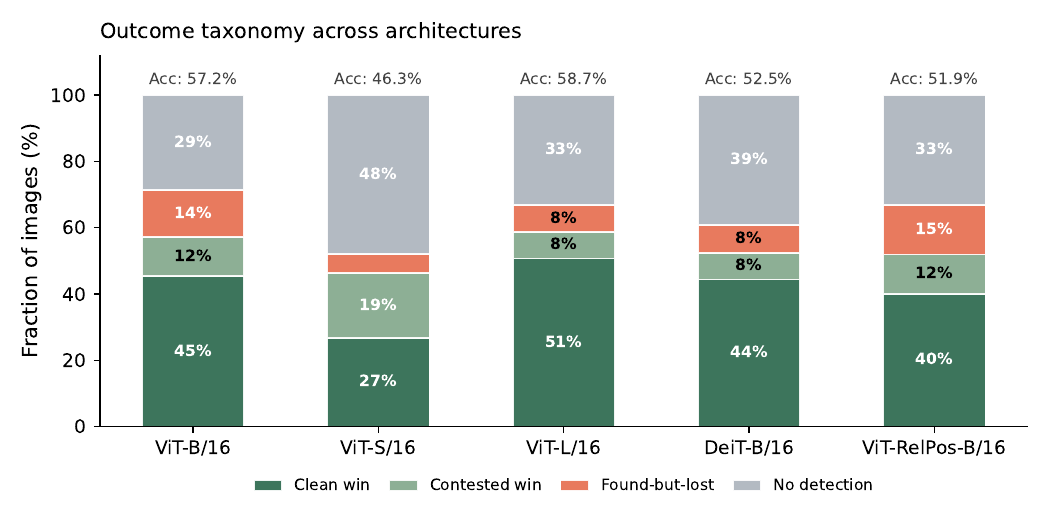}
    \caption{\textbf{Outcome taxonomy across architectures.}
    Each bar shows the fraction of composite images assigned to each outcome category: clean win, contested win, found-but-lost (FBL), and no detection. Outcomes are defined using patch-level predictions at post-norm as described in Appendix~\ref{sec:appendix_taxonomy}. Baseline accuracy is shown above each bar. ViT-S/16 shows the highest no-detection rate and the lowest FBL rate, indicating that many of its errors reflect weak or absent object evidence rather than recoverable aggregation failure.}
    \label{fig:appendix_cross_model_taxonomy}
\end{figure}

We first compare the outcome taxonomy across architectures. Figure~\ref{fig:appendix_cross_model_taxonomy} shows that the same four outcome types appear across models, but their relative frequencies differ substantially. ViT-S/16 has the largest no-detection fraction and the smallest FBL fraction, indicating that many of its errors reflect weak or absent object evidence rather than recoverable aggregation failure. In contrast, ViT-B/16 and ViT-RelPos-B/16 show the largest FBL fractions, indicating a larger pool of errors where object evidence is present at the patch level but not selected by the final prediction.

These differences in error composition help explain why the recovery probe transfers unevenly across models. Models with larger FBL pools provide more opportunities for re-ranking to recover object evidence, whereas models dominated by no-detection errors leave less recoverable structure for the probe to exploit.

\FloatBarrier

\subsection{Patch Expansion Dynamics Across Architectures}
\label{app:patch_gen}

\begin{figure}[b!]
    \centering
    \includegraphics[width=\textwidth]{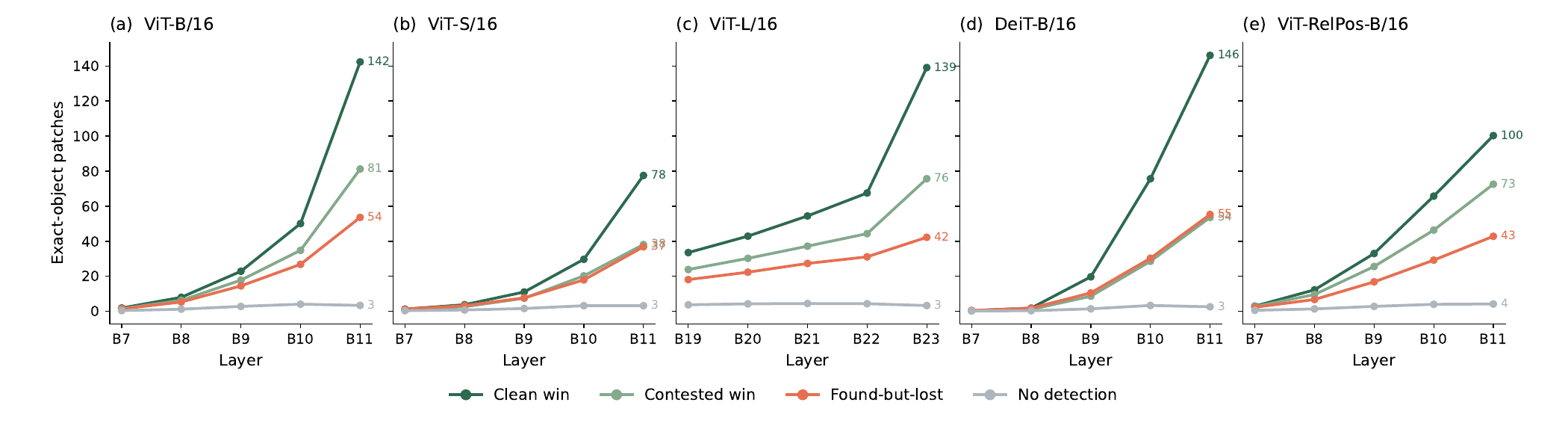}
    \caption{\textbf{Object patch expansion across ViT variants.}
    Each panel shows the mean number of patches, out of 196, whose argmax logit-lens prediction matches the object class, plotted across the final transformer blocks and post-norm, split by outcome category. Across architectures, clean wins show broad expansion, FBL cases stall at an intermediate level, and no-detection cases remain near baseline. This shared pattern supports the interpretation that FBL cases reflect incomplete propagation of object-aligned evidence rather than a failure unique to ViT-B/16.}
    \label{fig:appendix_cross_model_patch}
\end{figure}

We next ask whether FBL cases show similar spatial dynamics across architectures. Figure~\ref{fig:appendix_cross_model_patch} tracks the number of patches whose logit-lens prediction matches the object class across the final transformer blocks and post-norm. Across models, clean wins show broad expansion of object evidence across the patch grid, FBL cases stall at an intermediate level, and no-detection cases remain near baseline. This shared pattern suggests that FBL cases capture a similar failure mode across model capacity, training procedure, and pooling mechanism.

\FloatBarrier

\subsection{Near-Miss Structure and Fixed-Rule Recovery}
\label{app:fix_gen}

Across models, the central FBL signature is preserved. FBL cases remain near-miss errors, with the correct class appearing in the final top-5 predictions in 94.7--98.7\% of cases across models. For ViT-B/16, the correct class has a median rank of 2. We then apply the same re-ranking rule used in Table~\ref{tab:recovery_gates} to all models, using thresholds $a > 1.0$, $g < 1.0$, and $d > 0.5$ at the penultimate block, without architecture-specific retuning. This fixed rule allows recovery performance to be compared across architectures without model-specific tuning.

Table~\ref{tab:cross_model_generalization} shows that the fixed rule yields net-positive accuracy changes in all models, although the recovery-to-breakage tradeoff varies substantially. Net $\Delta$ is computed over all images and therefore includes incidental corrections of non-FBL errors, whereas the recovery-to-breakage ratio refers specifically to FBL cases recovered relative to correct predictions broken. The rule transfers best to models with stronger spatial compactness signals and larger FBL pools, and less well to ViT-S/16, where FBL cases are rare and no-detection errors dominate. Descriptively, models with larger compactness gaps broadly show larger recoverable fractions, but we treat this association as descriptive rather than inferential because the comparison includes only five architectures.

\begin{table}[t!]
\centering
\caption{\textbf{Cross-architecture robustness of aggregation-failure diagnostics.}
For each model, we report accuracy on the stratified composite set, FBL rate, the fraction of FBL cases where the correct class appears in the final top-5 predictions, the compactness gap measured at the penultimate block, and recovery performance under the fixed re-ranking rule from Table~\ref{tab:recovery_gates}. The same rule, $a > 1.0$, $g < 1.0$, and $d > 0.5$, is applied uniformly across all models without architecture-specific retuning. ``Net $\Delta$'' is computed over all images and includes incidental corrections of non-FBL errors. ``FBL ratio'' and ``FBL rec.'' are computed only with respect to FBL recovery.}
\label{tab:cross_model_generalization}
\small
\begin{tabular}{lcccccccc}
\toprule
\textbf{Model} & \textbf{Pool} & \textbf{Acc} & \textbf{FBL} & \textbf{Top-5} & \textbf{Gap} & \textbf{Net $\Delta$} & \textbf{FBL ratio} & \textbf{FBL rec.} \\
\midrule
ViT-B/16   & token & 57.2\% & 14.1\% & 98.7\% & 49\% & +5.7pp & 9.4:1 & 40.6\% \\
ViT-S/16   & token & 46.3\% &  5.7\% & 94.7\% & 17\% & +0.9pp & 0.7:1 & 13.8\% \\
ViT-L/16   & token & 58.7\% &  8.2\% & 96.2\% & 20\% & +1.1pp & 2.3:1 & 16.4\% \\
DeiT-B/16  & token & 52.5\% &  8.3\% & 96.9\% & 22\% & +1.9pp & 2.2:1 & 23.9\% \\
ViT-RelPos-B/16 & avg & 51.9\% & 14.9\% & 95.1\% & 23\% & +1.9pp & 1.9:1 & 22.2\% \\
\bottomrule
\end{tabular}
\end{table}

Together, these results show that FBL errors are near-miss errors with consistent spatial dynamics across transformer variants. The ViT-RelPos result further indicates that the FBL signature is not specific to CLS-token pooling. Even when final predictions are obtained by averaging patch tokens, object-aligned evidence can remain spatially localized and be outweighed by competing background-aligned evidence. The strength of recovery varies across models and is larger when the spatial compactness signal is stronger, suggesting that recoverability depends on how clearly object evidence remains localized before final aggregation.